\documentclass{article}

\usepackage[english]{babel}
\usepackage{amssymb}
\usepackage{caption}
\usepackage{algorithm}
\usepackage{algorithmic}
\usepackage{tikz}
\usetikzlibrary{arrows.meta, shapes.multipart, shapes, arrows}
\usepackage{indentfirst}

\usepackage[letterpaper,top=2cm,bottom=2cm,left=3cm,right=3cm,marginparwidth=1.75cm]{geometry}

\usepackage{amsmath}
\usepackage{graphicx}
\usepackage{dsfont}
\usepackage[colorlinks=true, allcolors=blue]{hyperref}

\usepackage{authblk}

\newtheorem{remark}{Remark}
\newcommand{\scal}[1]{\left< #1 \right>}
\newcommand{\norm}[1]{\left\| #1 \right\|}
\newcommand{\argmin}{\mathrm{Argmin}}
\newcommand{\1}{\mathds{1}}

\newcommand{\paren}[1]{\left( #1 \right)}
\newcommand{\croch}[1]{\left[\, #1 \,\right]}
\newcommand{\acc}[1]{\left\{ #1 \right\}}

\newcommand{\comment}[1]{}

\title{Representation learning with a transformer\\
by contrastive learning\\ for money laundering detection}

\author[1, 2]{Harold Guéneau \thanks{harold.gueneau@proton.me}}
\author[1]{Alain Celisse}
\author[3]{Pascal Delange}
\affil[1]{Laboratoire SAMM\\
Université Paris 1 Panthéon-Sorbonne, Paris, France}
\affil[2]{Shine}
\affil[3]{Marble}

\begin{document}
\maketitle

\begin{abstract}
%
The present work tackles the money laundering detection problem. A new procedure is introduced which exploits structured time series of both qualitative and quantitative data by means of a transformer neural network.
The first step of this procedure aims at learning representations of time series through contrastive learning (without any labels).
The second step leverages these representations to generate a money laundering scoring of all observations.
A two-thresholds approach is then introduced, which ensures a controlled false-positive rate by means of the Benjamini-Hochberg (BH) procedure.
Experiments confirm that the transformer is able to produce general representations that succeed in exploiting money laundering patterns with minimal supervision from domain experts. 
It also illustrates the higher ability of the new procedure for detecting non-fraudsters as well as fraudsters, while keeping the false positive rate under control. This greatly contrasts with rule-based procedures or the ones based on LSTM architectures.

\end{abstract}

\section{Introduction}

%
Money laundering represents a significant global concern with an annual estimated amount from 2 to 5\% of the world GDP \cite{aei2103318}.
It strongly impacts the ability of governments to collect taxes and fight crime, damaging financial institutions' stability and affecting the economic development.
Therefore, addressing money laundering has become an important challenge for political and financial institutions to prevent these substantial financial losses.
The United Nations (UNODC) defines money laundering as the processing of criminal proceeds to disguise their illegal origin \cite{moneylaunderingunodc}, making its detection particularly challenging.
For instance perpetrators mimic legal financial behaviors by potentially keeping accounts dormant for years with an licit activity before using them fraudulently.
Another difficulty results from frequent changes of money laundering patterns along the time as public policies and other environmental factors make new fraud patterns more lucrative \cite{Lyeonov2024FINANCIALFA, Sinno2023TheEO}.
This evolving nature of money laundering makes their detection even more challenging and constantly requires new detection strategies since traditional ones can quickly become obsolete.

%
Financial institutions operate under strict regulatory requirements for money laundering detection and suspicious activity reporting. 
A non-compliance can result in a revocation of the financial license or large fines for any customer that would be proved as negligent. This falls under the catch-all concept of “anti-money laundering” framework (AML) \cite{AMLD6_2024}.
Financial institutions therefore employ dedicated teams of financial analysts in order to comply with these regulations and to avoid financial losses.
However due to the high volume of customers and transactions in modern financial systems, these investigations cannot be entirely hand-made.
%
In order to let the financial analysts handle only the last steps of investigation, financial institutions have implemented so-called automated "rule-based systems" \cite{Hossam2016, Laurenza2020}.
These are typically based on expert knowledge of past fraud patterns. For instance the trigger event can be a fraudulent payment, a specific customer activity, a periodic review of all accounts.
Rule-based systems provide explicit audits and have some explanatory capabilities, which is an asset in a regulatory context (banks are expected to provide explanations about each suspicious case, which can help the analysts during the investigation).
While rule-based systems provide a framework with inherent explainability, their efficiency suffers severe limitations in money laundering detection. Such rule-based detections often exhibit a very low precision with from 95\% to 98\% of false positives \cite{aei2103318}.
Such a deficiency of rule-based systems results from their high sensitive to hard-coded thresholds, which are difficult to update in real time.
Frequent modifications of these rules would be necessary as fraud patterns evolve, which turns out to be difficult due to the inertia of large financial institutions.


By contrast, machine learning approaches \cite{9446887} can address a lot of ongoing challenges, and complement existing rule-based frameworks, for instance by reducing the dependency on such scenarios or by uncovering new patterns associated with money laundering.
The most straightforward approach to design such a machine learning approach is to consider tabular data in input made of macroscopic features describing the customer and its associated activity (transaction aggregates, or graph descriptors for example) \cite{eddin2022antimoneylaunderingalertoptimization, 9446887, Sudjianto01022010, Tertychnyi2020ScalableAI}.

However a lot of information is lost when designing these aggregated features to reshape the available knowledge into tabular data. Moreover, designing meaningful aggregates also requires some expert knowledge, while being limited in the detection of new patterns.
By contrast, the present work addresses this question by considering the whole set of raw transactions (not only aggregates or other summary metrics) collected along the time (\cite{10.1007/978-3-031-34671-2_17} for example).
To be more specific, a transformer is trained from raw transaction time series in the same spirit as what is made in HAMLET \cite{10.1007/978-3-031-34671-2_17}, with some differences.  
However only one transformer is exploited in the present context for encoding the whole time series, unlike HAMLET which rather relies on two transformers (one for encoding the transactions and the other for the time series).
Transformer neural networks were initially developed in the context of Natural Language Processing. Since language can be interpreted as a "time series of words", a transformer can capture temporal dependencies across more general types of time series \cite{rodriguez2022natural,dalla2023nucleotide, kisiel2022portfoliotransformerattentionbasedasset, wen2023transformers}. 
%
In the present money laundering context, the transformer is learning representations from raw transaction streams, where each event is composed of both quantitative and categorical attributes. This creates a far larger "vocabulary" of possible feature combinations that represents diverse types of financial events.

Another important difference between Hamlet ( \cite{10.1007/978-3-031-34671-2_17}) and the present work owes to the use of contrastive learning as a means for (pre-)training the transformer without any labeled data, whereas HAMLET does exploit the available labels along the training process.
Indeed beyond input data structure, accurately detecting money laundering is challenging because of the lack of correctly labeled data. Fraudulent customers remain hard to identify, with incomplete descriptors, and the expected overall proportion of fraudulent transactions is (fortunately) low compared to legal transactions, which makes the learning problem highly imbalanced \cite{9446887}.
Contrastive learning (\cite{hu2024comprehensive}) and more generally self-supervised learning (\cite{gui2024survey}) have proven to be of great help in such situations where truly labeled data are not available. For instance \textit{Contrast to Divide: Self-Supervised Pre-Training for Learning with Noisy Labels} \cite{Zheltonozhskii_2022} demonstrates the advantage in considering a pre-training step in presence of noisy labels. Moreover, Self-Supervised Learning (SSL) methodologies (including contrastive learning) maintain performance under severe class imbalance \cite{liu2022selfsupervisedlearningrobustdataset}.
Contrastive learning has been already used with time series. For instance, Contrastive Predictive Coding \cite{oord2019representationlearningcontrastivepredictive} has been successfully applied to a variety of time series. However capturing the temporal relationships in the context of financial time series is of different nature compared to classical use-cases where Contrastive Predictive Coding has been used up to now (more similar to NLP problems).

\medskip

\begin{figure}[H]
  \centering
  \includegraphics[width=1\textwidth]{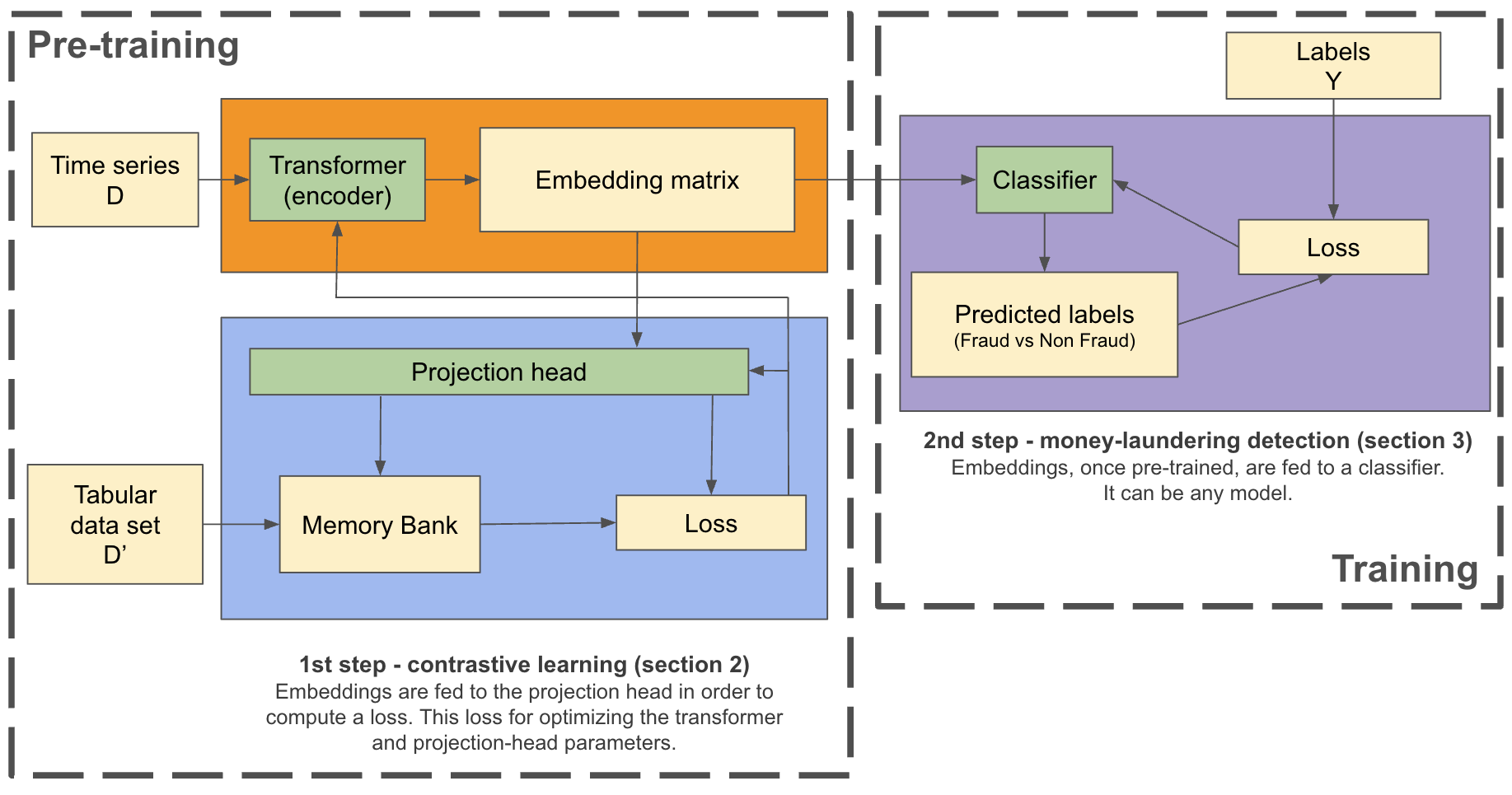}
  \caption{Summary of the proposed process}
  \label{fig:transformer_with_cl_overview}
\end{figure}

%
The main contributions of the present work are three-fold:
As illustrated by Figure~\ref{fig:transformer_with_cl_overview}, a first contribution consists in feeding a transformer architecture with raw time series of transactions instead of aggregates. This reduces the usual reliance on expert knowledge for designing such aggregates summarizing the valuable information. 
One specificity of the present context is that each transaction is described by a vector of both qualitative and quantitative descriptors, which makes the time series a complex structured object. 
%
A second contribution of the present work stems from the use of contrastive learning for (pre-)training the transformer without labeled data. 
 %
 By contrast with the supervised contrastive learning detailed in \cite{khosla2021supervisedcontrastivelearning}, contrastive learning is performed with no label.
 %
 %
 %
It rather leverages a similarity criterion which allows for deciding if two observations are similar or not. This refines the selection of positive and negative examples with more nuances, potentially capturing relationships that might be missed if relying only on class labels.
 %
 %
%
Our third main contribution to the specific field of money laundering detection relies on introducing a two-threshold classification procedure (see Figure~\ref{fig:transformer_with_cl_overview}) for distinguishing non-fraudulent accounts from fraudulent ones. Unlike most of ongoing rule-based strategies, these two thresholds are calibrated by means of the Benjamini-Hochberg multiple testing procedure, which allows for controlling the False Discovery rate (FDR) criterion and therefore avoids too many false positives.
All of this results in a considerably reduced amount of false negatives compared to state-of-the-art rule-based strategies.      

\medskip

The remaining part of the paper is structured as follows. Section~\ref{sec.representation.learning} details the specificities of the money laundering detection, and also describes how to learn the representation of our complex time series. The contrastive learning step (for the transformer pre-training) is exposed in Section~\ref{sec:transformer.contrastive.learning}, while the sampling of positive and negative examples is explained along Section~\ref{sec:positive.negative.example.sampling}. 
The application of the representation learning strategy to the money laundering problem is then explained in Section~\ref{sec.money.laundering}.
The two-thresholds strategy used for detecting fraudulent accounts and non-fraudulent ones is exposed in Section~\ref{sec.Two.thresholds}, while the calibration of these two thresholds by means of the BH procedure is detailed in Section~\ref{sec.thresholds.BH}. 
Section~\ref{sec.experiments} finally illustrates the general behavior of the full procedure on simulation experiments carried out from a (subset of a) real-life anonymized dataset. 
%

\section{Representation learning of financial time series}
\label{sec.representation.learning}

The present section describes the contrastive pre-training step, which corresponds to the blue box (first step) of Figure~\ref{fig:transformer_with_cl_overview}. The goal of this step to produce representations of time series of financial events (qualitative and quantitative descriptors). Such a representation embeds the financial time series into a finite-dimensional vector also named as an \emph{embedding}.
Sections~\ref{sec:dataset.description} and~\ref{sec:objectives.representations} introduce the structure of the dataset and the main elements of the pre-training step. 
Then Section~\ref{sec:transformer.contrastive.learning} details the contrastive learning procedure used to train the transformer neural networks (seen as an encoder).
As a key ingredient of the contrastive learning strategy, the sampling of negative and positive examples is detailed along Section~\ref{sec:positive.negative.example.sampling}.
Finally, Section~\ref{sec:related.works.in.representation.learning} discusses related works in representation learning.

\subsection{Data description and notation}
\label{sec:dataset.description}

\begin{figure}[H]
  \centering
  \includegraphics[width=0.75\textwidth]{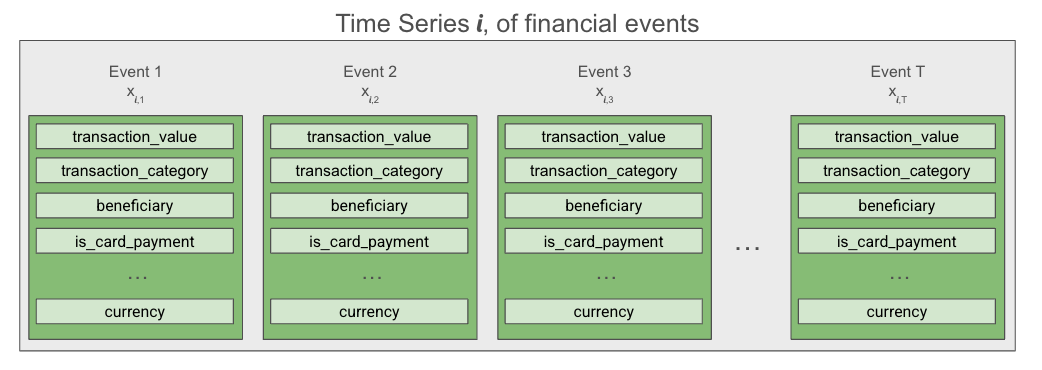}
  \caption{Illustration of a time series in the dataset $\mathcal{D}$}
  \label{fig:data.d}
\end{figure}

Let us consider a dataset $\mathcal{D}= \{ x_0, x_1, x_2, \ldots, x_{N-1} \}$ of $N$ observations. 
For each $1\leq i\leq N$, it is important to figure out that $x_i \in \mathbb{R}^{T \times d_{input}}$ denotes a time series of length $T$. Each time series $x_i = (x_{i,1},\dots,x_{i,T})$ is made of $T$ vectors of dimension $d_{input}$ describing $T$ successive financial events (transactions). Each vector $x_{i,t}$ has both quantitative and qualitative features describing the event (see Figure~\ref{fig:data.d}). 
These features can be the direction of money transfer, the type of transaction (e.g. card payment or bank transfer), the amount of transferred money, or the date for instance. Let us mention that qualitative data can be encoded through a categorical encoder such as the One-hot encoding \cite{zhu2024comparativestudyperformancecategorical} or the target encoding \cite{miccibarreca2001preprocessing} for example.

For motivating the use of a transformer \cite{vaswani17} to encode such a time series of events, let us mention that each of these financial events is very similar to a word within a sentence (with its specific ordering). Therefore similarly to what is usually made in NLP \cite{reimers2019sentencebertsentenceembeddingsusing}, a time series of such events can be embedded using a transformer deep neural network that will learn the financial semantic without using any label.

\bigskip

In addition to this dataset $\mathcal{D}$, each individual also comes with complementary descriptors so that one can define a new set $\mathcal{D^{'}}= \{ x^\prime_0, x^\prime_1, x^\prime_2, \ldots, x^\prime_{N-1} \}$, where every $x^\prime_i \in \mathbb{R}^{d_{comp}}$ is a vector of $d_{comp}$ numeric descriptors associated with the time series $x_i$.
\begin{figure}[H]
  \centering
  \includegraphics[width=0.75\textwidth]{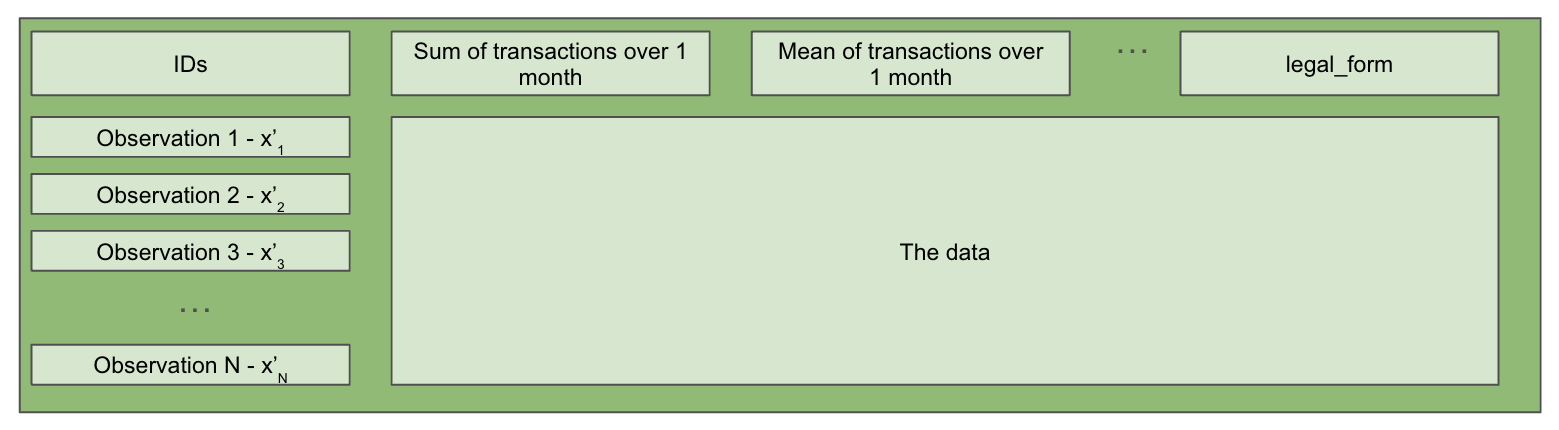}
  \caption{Illustration of the dataset $\mathcal{D}^\prime$}
  \label{fig:data.d.prime}
\end{figure}
As illustrated by Figure~\ref{fig:data.d.prime}, these features can be both categorical and numerical. The categorical ones include characteristics such as the legal form of a customer or the industry sector classification. The numerical features are mainly simple aggregated metrics derived from the time series in $\mathcal{D}$, it can be the mean or the sum of the averaged payout value over a period of time.
Let us emphasize that the vectors within $\mathcal{D^{'}}$ lack expressiveness to achieve optimal performance across a broad range of setups particularly in the context of money laundering detection. This claim is supported by the simulation experiments results described in Section~\ref{sec.learning.classif.threshold.experiments} where tabular data (aggregates) cannot achieve the same statistical performance as the raw time series.
However, the information in $\mathcal{D^{'}}$ can still be leveraged to design a similarity criterion and therefore decide which observations are similar, and which are not. This similarity criterion is a corner stone of the (unsupervised) contrastive learning strategy (see Section~\ref{sec:transformer.contrastive.learning}).

\medskip

In the money laundering detection context, the data are classically highly imbalanced since most of accounts are non-fraudulent ones, as described in Section~\ref{sec.Data.Description} where at most 5\% pf the test set data are fraudsters.
The proposed two-steps methodology as well as some aspects of the pre-training made by contrastive learning help in mitigating the strong imbalance in the data.

\subsection{General pipeline for representation learning: Pre-training}
\label{sec:objectives.representations}

\begin{figure}[H]
  \centering
  \includegraphics[width=1\textwidth]{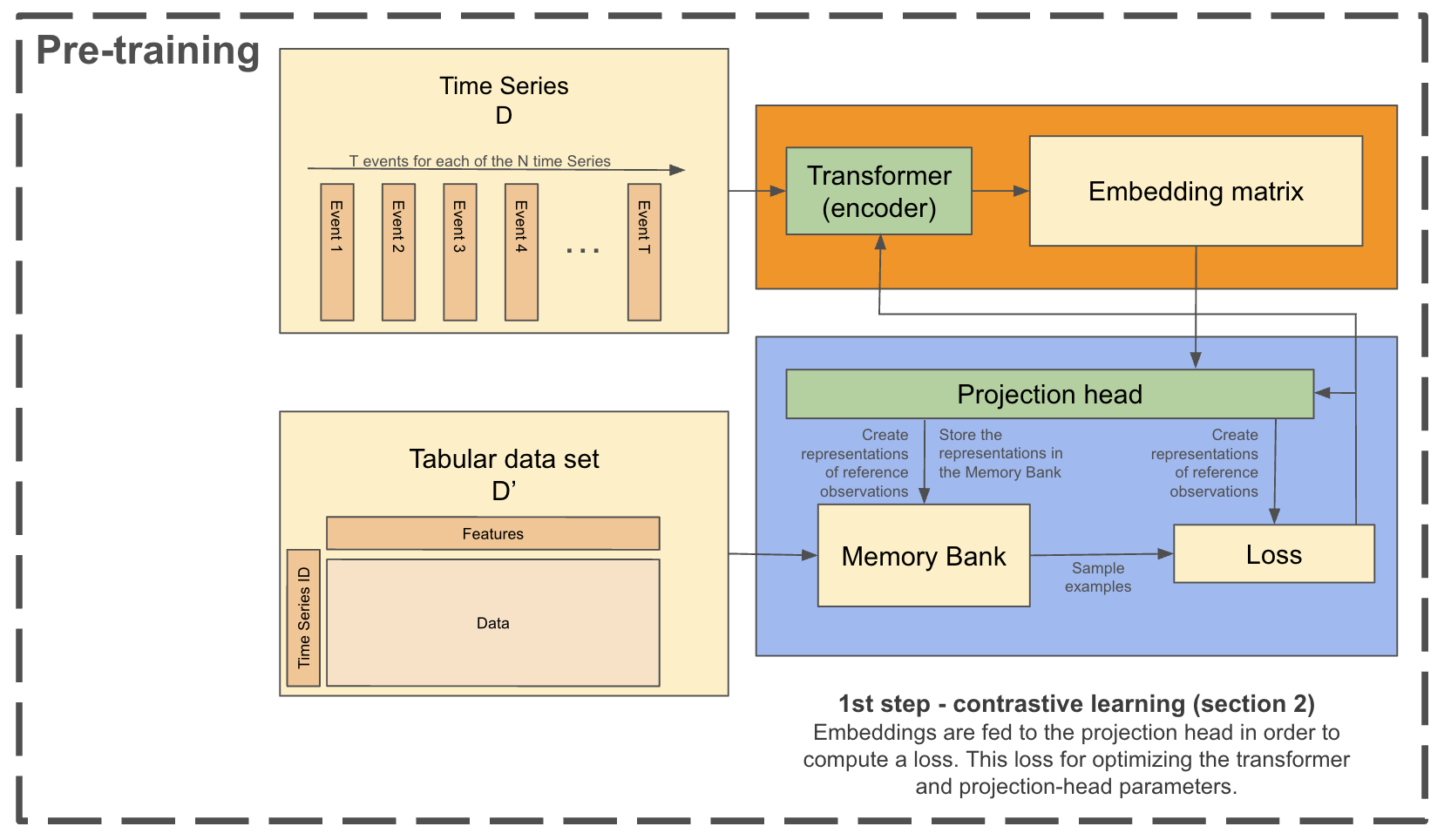}
  \caption{Contrastive pre-training of the transformer encoder}
  \label{fig:transformer_encoder_pre_training}
\end{figure}

 The general pipeline for representation learning relies on several blocs that are combined as illustrated by Figure~\ref{fig:transformer_encoder_pre_training}, which summarizes the whole pre-training process.
 \paragraph{Transformer bloc}
 The dataset $\mathcal{D}$ is used as an input of the transformer neural network (orange bloc). 
 The main merit of the transformer is to output an embedding matrix, which encodes the initial dataset $\mathcal{D}$ as $N$ row-vectors in a latent space of finite dimension.
 The transformer neural network architecture \cite{vaswani17} is a powerful tool for processing time series \cite{wen2023transformers}. It has proven great efficiency in a variety of fields, particularly in NLP, but also in biology \cite{dalla2023nucleotide} or in finance \cite{kisiel2022portfoliotransformerattentionbasedasset}.

\paragraph{Contrastive learning bloc}
The embedding output by the transformer is then sent in input of the contrastive learning step (blue bloc), the goal of which is the transformer pre-training.
Contrastive learning is a robust and flexible methodology which leverages the auxiliary dataset $\mathcal{D^{'}}$ and enables us overcoming the lack of truly labeled data.
The auxiliary dataset $\mathcal{D^{'}}$ serves for designing a similarity measure used along the contrastive learning process following the principle: "similar individuals should have close representations".
%
%
This is achieved by optimizing a loss with respect to the weights of both the encoder (transformer) and the so-called \emph{projection head}, which is a Multi-Layer perceptron (MLP) Deep Neural Network (DNN) with the output layer of low dimension (for reducing the computational costs) \cite{chen2020simpleframeworkcontrastivelearning}.

Let us emphasize that a so-called \emph{memory bank} \cite{he2020momentumcontrastunsupervisedvisual} has been also introduced for saving the computational resources.
As illustrated by the bottom-right bloc within Figure~\ref{fig:transformer_encoder_pre_training}, the tabular data from $\mathcal{D^{'}}$ are stored in a "memory bank" with a limited size, which is a means for only storing relevant observations (therefore limiting the memory costs).

In the following, Algorithm~\ref{algo.pipeline} provides a high-level description (in pseudo-code) of the main steps of the contrastive learning process that will be further explained along the next sections. 
%
The input of the contrastive learning are the weights of the transformer (encoder) denoted by the function $f_\theta$, and the ones of the projection head (dimension reduction) denoted by $h_\gamma$. The contrastive learning aims at optimizing all these weights simultaneously. The loss function used for this optimization is denoted by $\mathcal{L}_{\text{InfoNCE}}$ \cite{oord2019representationlearningcontrastivepredictive}.
\begin{algorithm}[H]
\caption{Pseudo code to illustrate contrastive pre-training of the transformer encoder}
\label{algo.pipeline}
\begin{algorithmic}[2]

\STATE \textbf{Input:} $\mathcal{D}$,  $\mathcal{D}^\prime$, number of epochs $\mathbf{e}$, number of batches $\mathbf{b}$

\STATE \textbf{Initialization:} $f_{\theta}(\cdot)$ (Transformer Encoder), $h_{\gamma}(\cdot)$ (Projection head), $\mathcal{L}_{\text{InfoNCE}}$ (Loss), $\mathcal{M} = \emptyset$ (Memory bank) 

\vspace*{2mm}
%
    

        
    
\FOR{$i \in [1, \mathbf{e}]  \cap \mathbb{N}$}
    \STATE Define a collection of $b$ batches of observations from $\mathcal{D}$: $$\mathcal{B}=\acc{B_1, \cdots, B_{\mathbf{b}}}$$
    \vspace*{-7mm}
    \FOR{$j \in [1, \mathbf{b}]  \cap \mathbb{N}$}
    
        \FOR{$ x_{ref} \in B_j $}
            \STATE Compute $$h_{\gamma} (f_{\theta} (x_{ref}))= h_{\gamma} (u_{ref})= z_{ref}$$ $\hfill \rightarrow$ Section~\ref{sec:transformer.contrastive.learning}
           \STATE Update $\mathcal{M}$: $$\mathcal{M} \leftarrow \mathcal{M}\cup \acc{z_{ref}} $$  $\hfill \rightarrow$ Section~\ref{sec:positive.negative.example.sampling}
           \\ 
           \STATE Sample one positive example and $K-1$ negative ones from $\mathcal{M}$:
           $$\mathcal{X} = \acc{z_+ } \cup \acc{z_-^1, z_-^2, \cdots, z_-^{K-1}} $$ 
            \STATE Sample noisy "copies" of examples from $\mathcal{X}$ ($\epsilon_+,\epsilon_i \sim \mathcal{N}(0, \sigma^2)$):
            $$\mathcal{X}^\epsilon = \acc{z_+ + \epsilon_+ } \cup \acc{z_-^1 + \epsilon_1, \cdots, z_-^{K-1} + \epsilon_{{K-1}}}$$ 
             $\hfill \rightarrow$ Section~\ref{sec:positive.negative.example.sampling}
            \STATE Compute the loss $\mathcal{L}_{\text{InfoNCE}}(z_{ref}, \mathcal{X}^\epsilon)$ $\hfill \rightarrow$ Section~\ref{sec:transformer.contrastive.learning}
        \ENDFOR
    \STATE Optimize $f_{\theta} (\cdot)$ and $h_{\gamma} (\cdot)$ (minimizing $\mathcal{L}_{\text{InfoNCE}}$) with stochastic gradient descent 
        
    \ENDFOR
    
\ENDFOR
\vspace*{2mm}
\STATE \textbf{Output} $f_{\theta} (\cdot)$
\end{algorithmic}
\end{algorithm}
At each step of the process, a \emph{reference} observations $x_{ref}$ is randomly chosen. The encoder function is composed with that of the projection head, which outputs a low-dimensional representation $z_{ref}$ that is sent to the memory bank.

The next step at the core of the contrastive learning process consists in a sampling step: Based on a similarity measure, a similar observation to reference one is chosen $z_+$ (called positive example) as well as dissimilar ones (called the negative examples) denoted $\acc{z_-^i}_{1\leq i\leq K}$.
As already suggested, the weights of the transformer combined with the projection head are optimized for maximizing the closeness between the representations of the reference observation and that of the positive examples while minimizing the distance between the representations of the reference example and the ones of the negative examples.



\subsection{Transformer and contrastive learning}
\label{sec:transformer.contrastive.learning}

In order to produce the required representation of complex time series, it is necessary to consider an encoder function $f_\theta$ (the transformer here) which maps the input (time series) $x_i \in \mathbb{R}^{T \times d_{input}}$, for $i \in [0, N-1]$ onto a finite-dimensional output vector $u_i$ (called the representation or the embedding) in a latent space $\mathcal{U}$ of dimension $d_{latent}$.
Along the contrastive learning (pre-training process), this representation in the latent space is "projected" onto a low-dimensional vector $z_i$ of dimension $d$ by means of the \emph{projection head} denoted by $h_\gamma$, defined from $\mathcal{U}$ to the $d$-dimensional space $\mathcal{Z}$, with $d\ll d_{latent}$ \cite{gupta2022understanding}. 

These two important steps are introduced and detailed in what follows.
%
%
\paragraph{Transformer}
The transformer architecture \cite{vaswani17} excels at capturing long-range dependencies, which can be crucial for highlighting long-range money laundering patterns. Such patterns can be operated over several months to be as discreet as possible.
One important feature of the transformer is the \emph{self-attention} mechanism \cite{vaswani17} it incorporates. This self-attention mechanism enables to focus on relevant long-range patterns while building the representations, especially with long sequences. 
Moreover compared to traditional time series encoders such as RNNs \cite{elman90, williams89} and LSTMs \cite{hochreiter97}, the transformer one has a lower computational complexity and is easier to parallelize (helping in processing large datasets) \cite{vaswani17}.


It is important to mention that the self-attention mechanism the transformer relies on is invariant by permutations. Without any positional information hard coded in the input observation, the transformer would not suitably account for the temporal structure of time series. Therefore, the transformer architectures require temporal (or positional) information to effectively process sequential data \cite{dufter2021positioninformationtransformersoverview}.
Unlike the original implementation by \cite{vaswani17} in the context of natural language processing (NLP), another specificity of the present work is to handle the sequential structure of transactions by learning the best positional embedding along the training process (\cite{wang2020positionembeddingslearnempirical}).
%

\begin{figure}[H]
  \centering
  \includegraphics[width=0.3\textwidth]{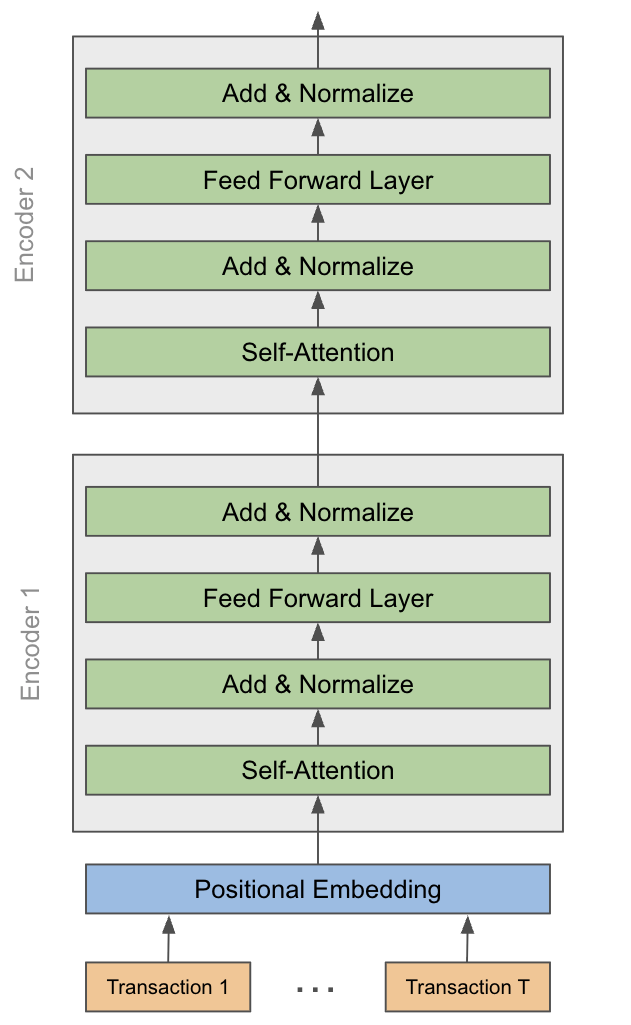}
  \caption{Illustration of the transformer encoder}
  \label{fig:transformer.architecture}
\end{figure}

The transformer (illustrated in Figure~\ref{fig:transformer.architecture}) performance is governed by a large set of hyperparameters such as the number of encoder layers, the latent space dimensionality, and the number of attention heads (\cite{vaswani17}). 
A few optimization parameters such as the learning rate and the dropout rate impact the convergence stability as well as the generalization performance. 
All the hyperparameters used along the experiments reported in Section~\ref{sec.experiments} are detailed in Appendix~\ref{appendix.hyperparameters.and.training.details}.

\paragraph{Motivating contrastive learning for money laundering detection}
One main idea of the present work is to train the transformer encoder by means of contrastive learning \cite{Le_Khac_2020, chen2020simpleframeworkcontrastivelearning, he2020momentumcontrastunsupervisedvisual}. The contrastive learning is a self-supervised learning (SSL) approach which further exploits (if available) the intrinsic data  structure to train a model without exploiting any label. In the present context this is made possible by designing a similarity criterion which allows for learning the semantics in the data \cite{Zheltonozhskii_2022}. 
This similarity can rely on the domain knowledge or on structural properties of the data. This methodology is highly relevant in the present context of money laundering detection, since labeled data are scarce or often unreliable.

\paragraph{Transformer and Projection head for contrastive learning}
The contrastive learning process integrates a projection head after the transformer encoder in order to improve the performance and the generalization power of the encoder \cite{xue2024investigatingbenefitsprojectionhead} \cite{gupta2022understandingimprovingroleprojection}.
This has been implemented in SimCLR for example \cite{chen2020simpleframeworkcontrastivelearning}. 
Along the contrastive learning process, the transformer encoder $f_\theta$ is composed with the projection head function $h_\gamma$ to output a final projected representation $z$ of an observation $x$. 
This architecture improves the model generalization using intermediate layers to extract more high-level and general features in $u$, while maintaining a task-specific optimization on another representation $z$, processed further in the neural network.
It is related to the logic studied in transfer learning (as described in \cite{transfer-learning-hee} for example), and to the idea of pre-training a large language model before fine-tuning it for a specific task in natural language processing \cite{howard2018universallanguagemodelfinetuning}.

More precisely, the encoding process starts with the transformer encoder $f_{\theta}(\cdot)$ with parameters $\theta$ (the transformer weights), generating a general-purpose representation $u \in\mathcal{U}$, with dimension $d_{latent}$ in a latent space. The purpose of this representation is to encode some valuable information that is hidden in the input space, which would improve the generalization performance.
This representation $u$ is then mapped into a smaller and specialized space through the projection head $h_{\gamma}(\cdot)$, with parameters $\gamma$ (the projection head weights). This projection head is classically an MLP neural network \cite{gupta2022understanding} parametrized by the number of hidden layers, the hidden‐layers dimensionality, the activation functions, a normalization strategy and a dropout rate. 
The choice of the hyperparameter values used in our experiments is detailed in Appendix~\ref{appendix.hyperparameters.and.training.details}.
The projection head map $h_\gamma(\cdot):$ $\mathcal{U} \to \mathcal{Z}$ produces a specific (projected) representation $z \in\mathcal{Z}$ such that $h_\gamma(u)=z$, with low dimension $d\ll d_{latent}$.
%
Let us also mention that the projection head reduces the optimization time complexity by reducing the dimension of the general representation $u$ of into a smaller space. The InfoNCE is indeed computed from $z$, which has a smaller dimension than $u$.

\paragraph{Sampling similar and dissimilar observations}
The present work combines time-series observations in $\mathcal{D}$ with tabular information in $\mathcal{D}^\prime$ used to design the similarity criterion (which is a key ingredient in contrastive learning).
For a so-called \emph{reference observation}, the tabular data from $\mathcal{D}^\prime$ are used to select \emph{positive} and \emph{negative examples} (related to the reference observation). Indeed, the set $\mathcal{D}^\prime$ contains aggregated metrics and descriptors used to measure similarity between candidate observations and the reference ones. Given a reference observation, the similarity criterion defined on $\mathcal{D}^\prime$ leads to distinguish between observations similar to the reference one (the so-called positive examples) and dissimilar ones (the negative examples).

Let $x_{ref}$ be the reference observation fr which an observation must be computed, and let $\mathcal{X}$ denote a set of $K$ example observations, with $x_{ref} \notin \mathcal{X}$, one positive example $x_+$ (expected to be similar to the reference observation $x_{ref}$), and $K-1$ negative observations (expected to be dissimilar to $x_{ref}$).
These observations are all processed through an encoder (composition of a transformer $f_\theta$ with a so-called projection head $h_\gamma$) to create an embedding of each observation.
Let $z$ be such an embedding of an observation $x$, $z_{ref}$ the one of a reference observation $x_{ref}$, and $z_+$ the one of a positive example $x_+$.
Then, one can define the set of positive examples $\mathcal{Z}_+$ and the one of negative examples $\mathcal{Z}_-$ by
\begin{align*}
& \mathcal{Z}_+ = \mathcal{Z}_+(z_{ref})  = \acc{ z \mid \mbox{$z$ is a positive example with respect to $z_{ref}$}  } \\
& z_+ = z_+(z_{ref}) \in \mathcal{Z}_+ \\
& \mathcal{Z}_- = \mathcal{Z}_-(z_{ref})  = \acc{ z \mid \mbox{$z$ is a negative example with respect to $z_{ref}$}  } \\
& z_- = z_-(z_{ref}) \in \mathcal{Z}_- \\
& \mathcal{X} = \acc{z_+} \cup \acc{z_-^1, z_-^2, \cdots, z_-^{K-1}}, \mbox{ with } z_+ \in \mathcal{Z}_+, \mbox{ and }\acc{z_-^1, z_-^2, \cdots, z_-^{K-1}} \in Z_-
\end{align*}
Notice that the notation $z_+=z_+(z_ref)$ for instance reflects the dependence of positive examples with respect to the reference observation that has been previously selected. The same remark also holds for negative examples, which therefore implies the same dependence for the set $\mathcal{X}$. Let us recall that this set is used along the minimization of the InfoNCE loss.

\paragraph{Increasing the diversity: Data augmentation by adding a Gaussian noise} 
According to Algorithm~\ref{algo.pipeline}, each positive and negative example is perturbed by adding a Gaussian noise $\mathcal{N}(0,\sigma^2)$, with $\sigma^2>0$. This mechanism helps in creating "new observations" from the original ones by Gaussian perturbation \cite{devries2017datasetaugmentationfeaturespace}, which yields realistic observations provided $\sigma^2$ remains small enough compared to the local structure of the data. 
In our experiments from Section~\ref{sec.experiments}, the Gaussian noise variance $\sigma^2$ is a hyperparameter the value of which is calibrated through a grid search strategy (see Appendix~\ref{appendix.hyperparameters.and.training.details}). 
Let us also emphasize that this data augmentation mechanism is applied at the level of the projection space of dimension $d$ that is, on finite-dimensional vectors $z_i=h_\gamma\circ f_\theta(x_i)$, for $1\leq i\leq N$. This contrasts with data augmentation strategies classically involved in contrastive learning where the observations are perturbed at the level of the original space (here the time series). Actually there already exist meaningful strategies for perturbing images in the original space \cite{chen2020simpleframeworkcontrastivelearning}. However to the best of our knowledge, practical analogues for time series do not exist yet, one possible explanation being the difficulty of preserving complex dependence structures along the time.

\paragraph{The InfoNCE loss}
The pre-training of the transformer (made by contrastive learning) relies on the optimization of the InfoNCE loss \cite{oord2019representationlearningcontrastivepredictive}.
This loss relies on a similarity measure $\rho(\cdot,\cdot):$ $\mathcal{Z}\times\mathcal{Z} \to \mathbb{R}_+ $. The choice made here for $\rho$ is the classical cosine similarity \cite{3320} given by
\begin{align*}
    \rho\paren{z,z^\prime} = \frac{\scal{z,z^\prime}}{\norm{z}\cdot \norm{z^\prime}},
\end{align*}
where $\scal{\cdot,\cdot}$ (respectively $\norm{\cdot}$) denotes the Euclidean scalar product (resp. the Euclidean norm) in $\mathcal{Z}$.

For a selected reference observation $x_{ref}$ (mapped through the projection head composed with the encoder) onto $z_{ref} = h_\gamma\circ f_\theta(x_{ref})$, positive and negative examples are randomly chosen to build the set $\mathcal{X}$ of cardinality $K$ as explained above.
The InfoNCE loss is then given by
\begin{align}
\label{eq_info_nce}
    \mathcal{L}_{\text{InfoNCE}} (\eta) = \mathbb{E} 
    \croch{- \log \left( \frac{ \exp( \rho(z_+, z_{ref}) / \mathcal{T} )}
    {\sum_{z \in \mathcal{X},z\neq z_+} \exp( \rho(z, z_{ref}) / \mathcal{T} )} 
    \right) },
\end{align}
where $\mathcal{T}>0$ denotes the so-called temperature parameter, and the expectation is computed over all possible choices of reference observation and related positive and negative examples. Let us also emphasize that $\eta=\paren{\theta,\gamma}$ is a parameter concatenating both $\theta$ and $\gamma$. The dependence of the InfoNCE loss with respect to $\eta$ reflects the fact that the reference vector $z_{ref}$ as well as all positive and negative examples within $\mathcal{X}$ depend on the weights of the transformer (denoted by $\theta$) and the projection head (denoted by $\gamma$). Therefore modifying any of these weights would result in a new final projection of these observations and then would vary the InfoNCE value.
For a reference observation, the weights $\theta$ of the transformer $f_\theta$ and $\gamma$ of the projection head $h_\gamma$ are optimized so that the computed representation of positive examples remains close to that of the reference observation, whereas the representation of negative examples is far away from that of the reference observation at the level of the projected space $\mathcal{Z}$ of dimension $d$.

The InfoNCE can be interpreted as a softmax on the similarity between the (projected) reference observation and the (projected) examples. 
The temperature $\mathcal{T}$ is an essential hyperparameter that drives the minimization. It aims at scaling the output of the softmax. 
Lowering the value of $\mathcal{T}$ results in requiring closer (in terms of similarity) representations for similar observations. Otherwise increasing the temperature allows for spreading the projected representation of similar observations.
In the experiments reported in Section~\ref{sec.experiments}, the temperature value is calibrated via a grid search strategy as explained in Appendix~\ref{appendix.hyperparameters.and.training.details}.
%

\subsection{Positive and negative examples selection}
\label{sec:positive.negative.example.sampling}

As explained along previous Section~\ref{sec:transformer.contrastive.learning}, the contrastive learning uses randomly selected observations as references and for each of them, then samples one positive example and $K-1$ negative examples relatively to this reference observation. 

The key ingredient behind this process is a similarity criterion defined from the complementary descriptors that have been gathered in the dataset $\mathcal{D}^\prime$.
In our context $\mathcal{D}^\prime$ contains a lot of information about the considered accounts: the average payout value over a period of time, and categorical features such as the legal form of the customer or for example the industry sector it belongs to. 
Therefore an observation $x_i$ is described within $\mathcal{D}^\prime$ in terms of descriptors (aggregates) collected within a vector $x_i^\prime \in\mathbb{R}^{d^\prime}$. For computing how similar two time series $x_i$ and $x_j$ are, the present approach introduces a similarity between vectors $x^\prime_i$ and $x^\prime_j$. In a nutshell, the respective representations of $x_i$ and $x_j$ must become closer as $x^\prime_i$ and $x^\prime_j$ are more similar to each other.

One idea presented in this work is therefore to sample positive and negative examples in $\mathcal{D}$, based on the similarity measured between elements in $\mathcal{D}^\prime$.
From an implementation perspective, a memory bank manages the sampling of these examples by storing the representations produced by the encoder $h_{\gamma} \circ f_\theta (\cdot)$, providing several key advantages as described below.

\paragraph{Memory bank}
This work implements a memory bank $\mathcal{M}$ that stores the embeddings produced during the training procedure. 
The memory bank is a tool already used in contrastive learning (see MoCo \cite{he2020momentumcontrastunsupervisedvisual} for instance).

Let us define $\mathcal{M}_b$ the memory bank at the iteration $b$ (i.e. after $b$ batches using the notations of Algorithm~\ref{algo.pipeline}).
After first initializing the memory bank with $\mathcal{M}_0 = \emptyset$, the memory bank is then updated as follows from each new batch of (projected) representations $\acc{z_1, \cdots, z_{\mathbf{b}_{size}}}$:
\begin{align}
\label{eq:memory_bank_update}
    \begin{cases}
        \mathcal{M}_{t+1} = \mathcal{M}_t \cup \acc{z_1, \dots, z_{\mathbf{b}_{size}}}, & \text{if } \vert \mathcal{M}_{t} \vert + \mathbf{b}_{size} < M\\
        \mathcal{M}_{t+1} = \paren{ \mathcal{M}_t  \setminus \mathcal{O}_t } \cup \acc{z_1, \dots, z_{\mathbf{b}_{size}}}, & \text{otherwise},
    \end{cases}
\end{align}
where $\mathcal{O}_t$ denotes the set of the $|\mathcal{M}_t| + \mathbf{b}_{size} - M$ oldest elements within $\mathcal{M}_t$, with $M$ denoting the maximum cardinality of the memory bank, and $\mathbf{b}_{size}$ being the current batch cardinality. In other words, the memory bank is iteratively updated by including the most recent elements, which results in removing the oldest elements from the memory bank each time the maximum cardinality $M$ has been achieved for the memory bank.

A memory bank also improves upon the selection of positive and negative examples by allowing more diversity in the sampling. Indeed, it stores representations over several batches ($\mathbf{b}_{size}<M$), which helps for reducing the dependency on a specific batch (especially when the batch size is small). 
Furthermore along the contrastive learning process, the memory bank stores the (projected) representations produced by different versions of the encoder combined with the projection head (the weights of which evolve along the iterations). 
As a matter of fact, this also helps stabilizing the training. On the one hand, it introduces a temporal ensembling \cite{laine2017temporalensemblingsemisupervisedlearning} by forcing the encoder to learn representations that remain consistent over the time (similarly to a momentum). On the other hand, the outdated embeddings produced by previous versions act as regularizers, adding some noise and then preventing from some overfitting.

\paragraph{Positive example sampling} 
For each reference observation $x_{ref} \in \mathcal{D}$ sampled from a batch, one positive example $x_+$ (with the corresponding representation $z_+ \in \mathcal{Z}_+$ associated) is required in order to compute the InfoNCE loss $\mathcal{L}_{\text{InfoNCE}} (\eta)$ (see Eq.~\ref{eq_info_nce}).
This positive example has to be hghly similar to the reference observation.

The memory bank $\mathcal{M}$ stores all the embeddings already processed among which one can choose a positive example. The present proposal builds positive examples by randomly sampling an observation from the top $\kappa$-closest neighbors in $\mathcal{D}^\prime$ ($\kappa\in\mathbb{N}^*$) to the reference observation, where the $\kappa$ neighbors have been ranked according to the similarity criterion over $\mathcal{D}^\prime$. In the present context, the similarity is the Euclidean norm in $\mathcal{D}^\prime$. 
$\kappa$ is a hyperparameter and needs to be calibrated for the considered dataset. In the experiments, $\kappa$ is set to 50 (the 50 closest companies with respect to the reference observation, as explained in Appendix~\ref{appendix.hyperparameters.and.training.details}).
The implementation relies on the approximate nearest-neighbor search \cite{10.1145/293347.293348} for saving the time and memory resources.

All of this is illustrated by Algorithm~\ref{alg.positive.example.sampling} in pseudo-code, which outputs an instance of a positive example given a memory bank $\mathcal{M}$ and a reference observation $x_{ref}$ (and its counterpart $x_{ref}^\prime$ within $\mathcal{D}^\prime$).
\begin{algorithm}[H]
\caption{Pseudo-code for positive example sampling}
\label{alg.positive.example.sampling}
\begin{algorithmic}[2]
\STATE \textbf{Input:} Memory bank $\mathcal{M}$, dataset $\mathcal{D}^\prime$, reference observation $x'_{ref} \in \mathcal{D}^\prime$, Number of neighbors $\kappa>0$

\vspace{2mm}
\STATE Compute the distance $r_\kappa$ between $x'_{ref}$ and its $\kappa$th neghbor in $\mathcal{D}^\prime\cap \mathcal{M}$
\STATE Set of $\kappa$-closest neighbors in $\mathcal{M}$: 
\begin{align*}
\mathcal{V}_\kappa(x^\prime_{ref}) = \acc{ 1\leq i\leq N\mid z_i\in\mathcal{M},\mbox{ and } \norm{ x^\prime_i - x^\prime_{ref}} \leq r_\kappa}
\end{align*}
\STATE Sample uniformly at random one index $i_+$ among $\mathcal{V}_\kappa(x^\prime_{ref})$
\STATE Positive example: $z_+ \leftarrow z_{i_+} \in \mathcal{M}$
\vspace{2mm}
\STATE \textbf{Output:} $z_+$
\end{algorithmic}
\end{algorithm}
The algorithm starts by computing the distance (Euclidean norm in $\mathcal{D}^\prime$) between $x_{ref}^\prime$ and its $\kappa$th neighbor in the memory bank. 
Let us mention that there is a slight abuse of notation here since the elements of the memory bank are the (projected) representations of the $x_i$s in $\mathcal{Z}$. Then $\mathcal{D}^\prime$ cannot formally intersect $\mathcal{M}$. Of course when writing this, one identify any vector $x^\prime_j$ with its representation $z_j$ in $\mathcal{Z}$ since the underlying observation $j$ is the same.  
As long as the distance $r_\kappa>0$ to the $kappa$th neighbor is known, then the set $\mathcal{V}_\kappa(x^\prime_{ref})$ of the $\kappa$-closest neighbors can be (uniquely) computed (if one moreover assumes that there is no tie). The final positive example is nothing bu any of the elements within $\mathcal{V}_\kappa(x^\prime_{ref})$ that has been chosen uniformly at random. Let us emphasize that any alternative strategy for choosing the positive example would be possible. For instance any importance sampling strategy taking into account the distance to $x_{ref}^\prime$ would make sense as well, although this is not the scheme that has been considered here.

\paragraph{Negative example sampling} 
Similarly, for each reference observation $x_{ref} \in \mathcal{D}$, a set $\mathcal{Z}_- \subset \mathcal{}Z$ of negative examples $z_-$ is designed (by considering the (projected) representations of the original observations $x$).

One known issue with negative examples is the risk of sampling among observations that are similar to the reference one \cite{huynh2022boostingcontrastiveselfsupervisedlearning}.
Committing such a mistake would mean that the optimization step would push away representations of similar observations, which is precisely what one tries to avoid.

An important contribution of the present work consists in a new strategy for selecting the negative examples based on a preliminary clustering of the observations within $\mathcal{D}^\prime$.
Indeed as long as $K\geq 1$ clusters have been computed from the similarity criterion over $\mathcal{D}^\prime$, then $x^\prime_{ref}$ does belong to a unique cluster. Therefore choosing any negative example among the observations belonging to the other $K-1$ remaining clusters is a means for avoiding the selection of observations that are too similar to the reference one.
The present proposal suggests to define the clustering by means of the Lloyd's algorithm for solving the K-means problem \cite{IKOTUN2023178}. Here the number of clusters is chosen from the data by maximizing the silhouette score \cite{Rousseeuw}.
 

In what follows, Algorithm~\ref{alg.negative.example.sampling} details the main steps of the strategy introduced above.
From a clustering algorithm $\Omega_K(\cdot):$ $\mathcal{D}^\prime \to \acc{1,\dots,K}$ that outputs the cluster label $\ell(x^\prime)$ of an observation $x^\prime\in\mathcal{D}^\prime$, Algorithm~\ref{alg.negative.example.sampling} starts by computing the label $\ell_{ref} = \ell(x^\prime_{ref})$ of the reference observation $x^\prime_{ref}$.
\begin{algorithm}[H]
\caption{Pseudo-code for negative example sampling}
\label{alg.negative.example.sampling}
\begin{algorithmic}[2]
\STATE \textbf{Input:} Memory bank $\mathcal{M}$, dataset $\mathcal{D}^\prime$, reference observation $x'_{ref} \in \mathcal{D}^\prime$, number of clusters $K\geq 1$, clustering algorithm $\Omega_K(\cdot)$ on $\mathcal{D}^\prime$.

\vspace{\baselineskip}
\STATE $\ell_{ref} = \ell(x^\prime_{ref}) = \Omega_K(x^\prime_{ref})$

\STATE Compute the indices of candidate negative observations: 
\begin{align*}
    \mathcal{I}_- = \acc{1 \leq i\leq N\mid \Omega_K(x^\prime_i) \neq \ell_{ref},\mbox{ and } z_i \in\mathcal{M} } 
\end{align*}
\STATE Sample randomly without replacement $K-1$ elements among $\mathcal{I_-}$: $$ \mathcal{S}_- = \acc{i_1,\ldots,i_{K-1}}$$  

\STATE Define the $K-1$ negative examples:
\begin{align*}
    \mathcal{Z}_- = \acc{z_-^1, z_-^2, \cdots, z_-^{K-1}} 
    = \acc{ z_j \in\mathcal{M} \mid j \in \mathcal{S}_-}
\end{align*}
\vspace{-5mm}
\STATE \textbf{Output:} $\mathcal{Z}_-$
\end{algorithmic}
\end{algorithm}
All indices of the observations that are not assigned to cluster $\ell_{ref}$ are collected within the set of candidate negative observations $\mathcal{I}_-$.
After randomly sampling $K-1$ distinct indices from $\mathcal{I}_-$, the set of negative examples is computed.
Any (refined) sampling scheme could be exchanged with the uniform without replacement one that is used along Algorithm~\ref{alg.negative.example.sampling}. For instance, keeping track of the distance of the candidate negative observations to $x^\prime_{ref}$ could be explored as well.

\paragraph{Gaussian noise injection}
Coming back to Algorithm~\ref{algo.pipeline}, the sampling of positive and negative examples is modified by adding some Gaussian noise, which results in substituting $\mathcal{X}$ with $\mathcal{X}^\epsilon$.

Adding such a Gaussian noise can be motivated by the need for exploring new situations (not covered within the dataset).  
Indeed covering all possible transaction events and patterns is not possible with a limited amount of data, which would limit the learning potential of the transformer.
This challenge is particularly relevant in the money laundering context where the labels are highly imbalanced with fraudulent patterns that are severely underrepresented.
It makes the task of learning the fraudsters' distribution a very challenging one.
In this respect, adding some Gaussian noise to the positive and negative examples allows for exploring new patterns.

Let us also mention that one specificity of the present proposal is to perturb the observations at the level of latent (projection) space $\mathcal{Z}$. This originality results from the difficulty of defining meaningful perturbations of time series which would respect the dependence structure along the time.
Notice also that in the present context, the variance $\sigma^2$ of the Gaussian noise is a means for controlling the level of perturbations added to the original observations. Another possible interpretation of the role of adding this Gaussian noise is to perform some regularization at the level of the latent space. This prevents the pre-training (learning the representation) to overfit the limited number of available observations.  
In the context of classification for example \cite{Bishop1995, NIPS2017_217e342f} reported that this helps in building smoother classification frontiers.

\begin{figure}[H]
  \centering
  \includegraphics[width=1\textwidth]{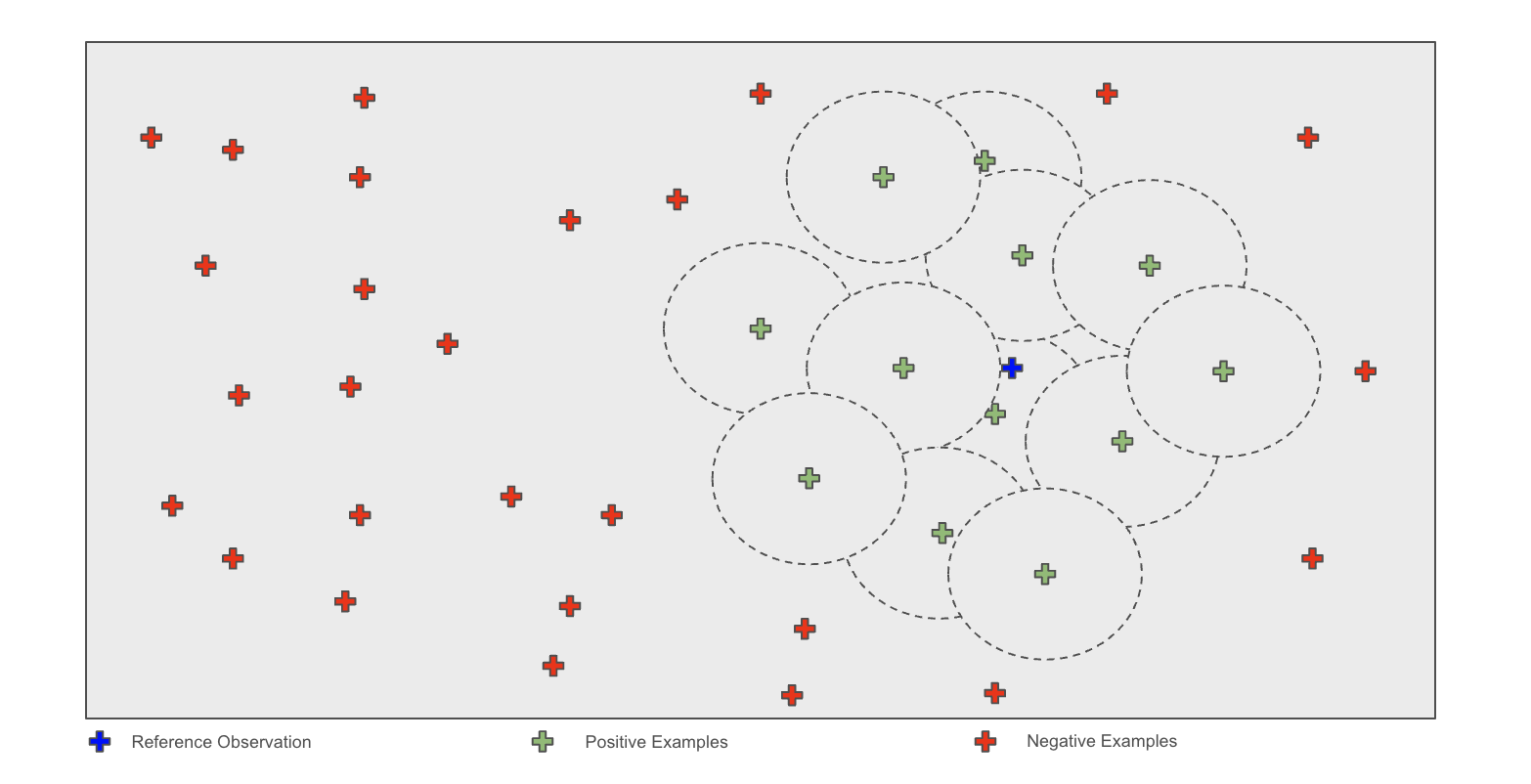}
  \caption{Illustration of how gaussian noise behave for positive example sampling}
  \label{fig:gaussian_noise_contrastive_learning}
\end{figure}

Presently Figure~\ref{fig:gaussian_noise_contrastive_learning} illustrates the idea underlying the addition of a Gaussian noise to the original data at the level of the latent space. 
For a reference observation (blue cross), the positive examples (green crosses) are similar to the reference observation. Therefore they should be close to the latter. According to our strategy (based on clustering) for defining negative examples (red crosses), these ones must to be "far away" from the reference observation (and at least further than positive examples). On the illustration, the dashed balls centered at the positive examples suggest the possible locations for Gaussian perturbed representation of any positive example $z_++\epsilon$. In particular, increasing $\sigma^2$ would result in increasing the radius of the balls.
All of this results in replacing the set $\mathcal{X}$ with $\mathcal{X}^\epsilon$ defined by  
\begin{align} \label{latent_space_noise}
    \mathcal{X}^\epsilon & = \acc{z_+ + \epsilon \cup \acc{z_-^1 + \epsilon_1, z_-^2 + \epsilon_2, \cdots, z_-^{K-1} + \epsilon_{K-1} } },
\end{align}
where $\epsilon_1,\ldots,\epsilon_{K-1} \sim \mathcal{N}_d(0,\sigma^2 I_d)$ are independent and identically distributed Gaussian random vectors.
Let us emphasize that such a perturbation strategy in the latent space can also serve to remedy the strong imbalance in the dataset between fraudsters and non-fraudsters (fraudsters being highly under-represented here). A similar idea has been already exploited in \cite{pmlr-v157-kim21a} with a different noise distribution.

\vspace{\baselineskip}
\subsection{Representation learning: Some related works}
\label{sec:related.works.in.representation.learning}
In Computer Vision (CV) for example, a classic approach of self-supervised learning could be to consider a convolutional autoencoder as explained by \cite{dong2017learningdeeprepresentationsusing}. 
The model is trained to compress and then reconstruct an image from an intermediate representation $z \in \mathbb{R}^d$ in a latent space. 
%
A similar strategy could be obtained by applying a LSTM autoencoder to time series~\cite{srivastava2016unsupervisedlearningvideorepresentations}. As above, this model is composed of a LSTM encoder compressing the time series in the latent space, and then of a second LSTM (decoder) that tries to reconstruct the original time series. This possible alternative is also explore in the present work as a benchmark compared to the use of a transformer (see Section~\ref{sec.distribution.FNF.experiments} for more details).

Regarding the self-supervised learning process, alternatives have already been discussed in the literature.
For instance, Natural Language Processing (NLP) can be seen as time-series modeling. Therefore it is common to approach text data with a transformer~\cite{vaswani17}. This transformer is pre-trained with self-supervised learning to capture the semantic of a language, mainly by addressing two main tasks:
One task consists in predicting the next word in a sentence (a GPT-like approach~\cite{Radford2018}). The other task rather randomly masks a word in a sentence and tries to retrieve it afterwards (a BERT-like approach~\cite{devlin19}).
These two strategies of self-supervised learning are suitable for language processing since they exploit the well-known sequential arrangement of the words. If one changes the order of some words in a sentence, the meaning can radically change: 
"The cat chased the mouse" and "The mouse chased the cat" share the same words while having two different meanings for example.
The words order in a sentence is a highly essential structure that has to be accounted for in NLP, which is somewhat different from what is considered in money laundering detection.
When dealing with transactions time series, the precise location of most of transactions will not impact the "meaning" of the resulting time series as much as it does in NLP. 
As a consequence, the rationale underlying a sequence of transactions is somewhat different from what is observed in NLP. This makes the use of the same semi-supervised learning approach (for instance trying to predict the next transaction) challenging in the context of money laundering.
In particular when considering a sequence of transactions, it is more relevant to tackle blocks of financial transactions in self-supervised learning.

\section{Money laundering detection}
\label{sec.money.laundering}
The main purpose here is to detail the process allowing to exploit the representations for money laundering detection designed from contrastive learning (see details in Section~\ref{sec.representation.learning}).
These representations are expected to be generic, and are meant to capture the semantic of financial transactions from the dataset $\mathcal{D}$.

Section~\ref{sec:from.representations.to.aml} introduces the downstream task of money laundering detection, and its connection with the learned representations.
As a by-product of these representations, a classification score can be computed that allows to classify any account as a fraudster or non-fraudster (binary classification). Section~\ref{sec.Two.thresholds} then describes a two-thresholds approach based on this score. This approach can help for investigating money laundering by removing the less suspicious observations, while having a focus on the most suspicious ones.
Finally, Section~\ref{sec.thresholds.BH} presents a procedure for calibrating the two thresholds to efficiently control the false positive rate.

\subsection{From representation learning to money laundering detection} \label{sec:representation_to}
\label{sec:from.representations.to.aml}

The approach described along Section~\ref{sec.representation.learning} mainly focuses on the representation learning that is, the pre-training of the transformer by contrastive learning.
The transformer (acting as an encoder) is pre-trained to produce a generic representation of the observations that extracts as much relevant information from the data (complex time series) as possible. An important feature is that \emph{this pre-training task does not require any labeled data.}
Once such a generic representation has been learned, it is then exploited to solve our downstream task, which is money laundering detection. 

By contrast with the pre-training task from Section~\ref{sec.representation.learning}, the specific money laundering detection comes with a few (uncertain) labeled data.
These labels result from a careful hand-made analysis of the accounts that have been flagged as ``potential anomalies" among which a few are labeled as fraudsters and the others are not. Let us also mention that the accounts that have not been raised as "potential anomalies" are considered as non-fraudsters by default. 

\begin{figure}[H]
  \centering
  \includegraphics[width=1\textwidth]{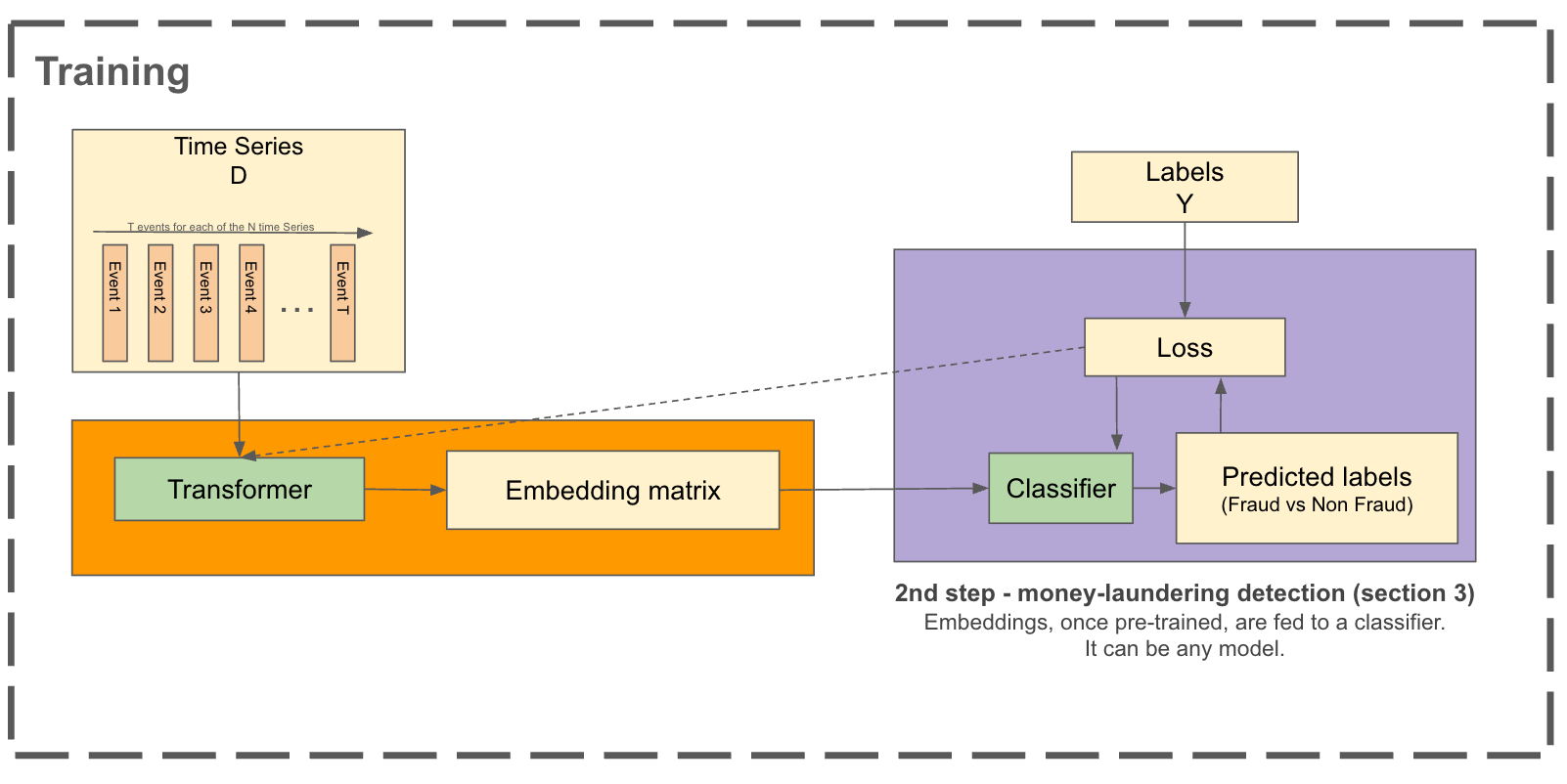}
  \caption{Classification process after contrastive pre-training}
  \label{fig:transformer_encoder_training}
\end{figure}

From now on, let $\mathcal{Y}= \acc{0,1}$ denote the set of possible labels. In the context of money laundering $y_i=0$ means the account $1\leq i\leq N$ is a "non-fraudster", whereas  $y_i=1$ means the account $i$ is a fraudster.
Let us further assume that the labels $y_i$ are given with the original descriptors $x_i$ so that one observes $N$ independent and identically distributed couples $(x_i,y_i)\in\mathcal{D}\times \mathcal{Y}$, for $1\leq i\leq N$. 
For addressing this new (classification) task, a classifier $g_\mu(\cdot):$ $\mathcal{U} \to \mathcal{Y}$ is now combined with the representation output by the transformer $f_\theta(\cdot)$. This combination is illustrated by Figure~\ref{fig:transformer_encoder_training}, while their classification performance at a point $(x,y)$ is measured by the classification loss $\ell_c(\cdot,\cdot):$ $\mathcal{Y}\times \mathcal{Y} \to \mathbb{R}_+$. In what follows, $\ell_c(\cdot,\cdot)$ is chosen to be the cross-entropy \cite{mao2023crossentropylossfunctionstheoretical}.
At this stage, the training can be performed in two ways: 
\begin{enumerate}
    \item The first one consists in considering the weights of the transformer as fixed (after the pre-training), and only optimize the classification loss with respect to the parameters $\mu$ of the classifier $g_\mu(\cdot)$. Therefore the training is that of the classifier given the representation, denoted by CR in what follows.
This leads to solve 
\begin{align} \label{eq_representations_as_features}
\mu^\star_{CR} \in \argmin_{\mu} \acc{\mathcal{L}_{classif-CR} (\mu)} ,
\end{align}
where $  \mathcal{L}_{classif-CR} (\mu)=\mathbb{E}_{(x,y)} \left[ \ell_c\paren{ y, g_{\mu} \paren{ f_{\theta} (x) } } \right]$.
In practice since $\mathcal{L}_{classif-CR} (\mu)$ depends on the unknown probability distribution of $(x,y)$, it is approximated using stochastic gradient descent (SGD) \cite{bottou2010large} with mini-batch (of cardinality 512) as suggested by Algorithm~\ref{algo.pipeline}. This outputs an estimator $\hat \mu_{CR}$ of the optimal classifier parameters (see Appendix~\ref{appendix.hyperparameters.and.training.details} for details about the choice of the hyperparameter values).  

    \item The second one consists in considering the weights $\theta$ of the transformer (output by the pre-training pipeline) as an "informed" initialization that can be further optimized. Then the alternative strategy jointly optimizes the classification loss with respect to both $\theta$ and $\mu$. This is called the transformer fine tuning, denoted by FT in Eq.~\eqref{eq_fine_tuning}.
the resulting criterion to be minimized is     
\begin{align} \label{eq_fine_tuning}
    \mathcal{L}_{classif-FT} (\theta, \mu) = \mathbb{E}_{(x,y)} \left[ \ell_c(y, g_{\mu} ( f_{\theta} (x) )) \right],
\end{align}
where the loss function is minimized with respect to $(\theta, \mu)$.
Following the same reasoning as above, the optimization is made by SGD, which results in an estimator $(\hat \theta_{FT}, \hat \mu_{FT})$ of the optimal weights.
\end{enumerate}
These two strategies are explored along Section~\ref{sec.distribution.FNF.experiments} where their practical performances are illustrated on simulation experiments from real data.

In what follows, $g_\mu$ has been chosen to be the logistic regression (binary) classifier \cite{https://doi.org/10.1111/j.1467-9574.1988.tb01237.x, Tertychnyi2020ScalableAI}. It is important to notice that, for each point $(x_i,y_i)$ ($1\leq i\leq n$), logistic regression also outputs a score $s_i$ that is an estimator of the \textit{a posteriori} probability for the individual $i$ to belong to class $1$ given its location $x_i$.
In Section~\ref{sec.Two.thresholds}, this score is exploited as an indicator of being a fraudster.
Moreover for improving the practical performance of the classifier, some $\ell_2$ regularization has been added on the unknown parameters for making the optimization problem strongly concave \cite{boyd2004convex}.

\subsection{A two-thresholds approach}
\label{sec.Two.thresholds}

Our money laundering detection approach can be conceptualized as a binary classification problem where a score between $0$ and $1$ is associated with each bank account. This score can be interpreted as the probability of being a fraudster given the observed location $x_i$.
As already mentioned, the prevalence of false positives in classical rule-based systems (\cite{aei2103318}) generates substantial operational costs and also limits the identification of high-risk money laundering cases.
In this respect Machine Learning can help in controlling this error rate at a prescribed level for avoiding any false positive inflation.

The classical money laundering context comes with a huge imbalance between the two classes to be recovered: the dataset only contains a very small proportion of fraudsters. This strong imbalance suggests to use a two-thresholds strategy for tackling differently non-fraudsters and fraudsters.
A first threshold is settled for removing the less suspicious observations from the list of observations triggered by alerts, and a second (larger) threshold is also designed for targeting the more suspicious observations for further investigations.

As a consequence, analysts can save time by only focusing on observations that are most likely fraudsters. On the other side, a lot of observations can now be removed from the set of suspicious observations, avoiding any waste of time by further investigating on them. 
\begin{remark}
At this stage, it should be clear that the above scores associated with all observations could serve for ranking the observations from the most suspicious one to the one with the lowest score.
The analyst could then focus on the top-$k$ scores for checking which ones are true positives among them. $k$ here is chosen according to the available time to spend on this step.
However the limitation of this approach is that there is no theoretical guarantee on the number of false positives induced by this naive strategy.
By contrast one originality of the present paper suggests two data-driven thresholds for detecting non-fraudsters and fraudsters while controlling relevant error rates at a prescribed level.
\end{remark}

\paragraph{Designing a low-threshold for discarding non-fraudsters} 
The aim is to detect a maximum number of non-fraudulent accounts while controlling the proportion of false positives. This first step allows for removing these accounts from the suspicious observations. This is achieved by setting a threshold $T_l$, called "low threshold", such that scores below this threshold are declared non-fraudulent. This low threshold $T_l$ is automatically chosen by controlling the proportion of mistakes among the list of suspicious accounts declared as non-fraudsters. The proportion of mistakes is quantified by means of the False Discovery Rate (FDR) criterion. \cite{https://doi.org/10.1111/j.2517-6161.1995.tb02031.x}.

\paragraph{Designing a high-threshold for detecting fraudsters}
The second step consists in focusing on the more risky observations which are likely to be fraudsters. Based on the intuition that the score should take large values for fraudsters, another threshold $T_h$, called "high threshold", is designed. The procedure declares as fraudsters all accounts with a score larger than this threshold. Based on a similar reasoning as the one exposed for the low threshold, the high threshold $T_h$ is automatically computed by maximizing the number of detections (accounts flagged as fraudsters by the procedure) while controlling FDR at a prescribed level. This control avoids any false positives inflation unlike what usually happens with classical rule-based strategies (\cite{aei2103318}).

\begin{figure}[H]
  \centering
  \includegraphics[width=0.75\textwidth]{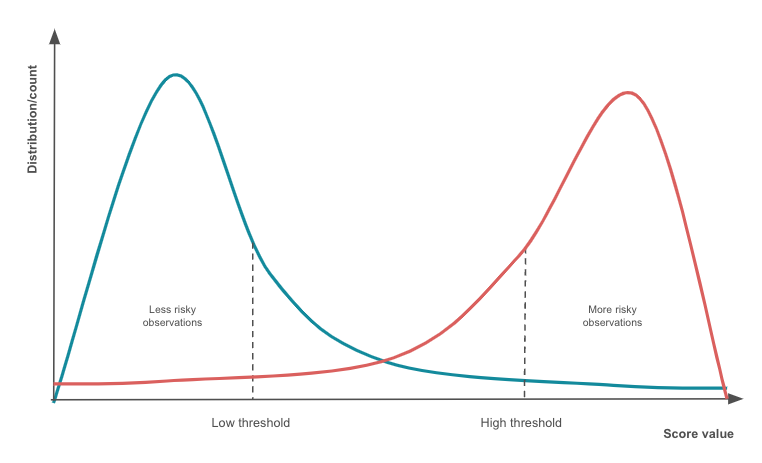}
  \caption{Illustration of the expected normalized distributions of the scores, conditionally of being non-fraudster (in blue) and of being fraudster (in red)}
  \label{fig:Score.distribution}
\end{figure}

As illustrated in the Figure~\ref{fig:Score.distribution}, the fraudulent scores in output of the model can be studied conditionally to the label (fraudulent accounts vs non-fraudulent accounts).
Fraudulent observations are expected to have scores on the far right side of the distributions supports, while the non-fraudulent observations should have scores on the far left side.
The two thresholds are helping in the analysts investigation by highlighting the more risky accounts and the less risky ones, at a prescribed FDR level. 
Let us insist on the fact that the FDR control is another originality of the present work in the context of money laundering detection compared to existing approaches in the domain.


\subsection{Threshold calibration via Benjamini-Hochberg procedure}
\label{sec.thresholds.BH}

The two-threshold approach briefly discussed in Section~\ref{sec.Two.thresholds} leads to considering two distinct classification problems: either rejecting accounts the scores of which are larger than $T_h$ (detected as fraudsters), or rejecting accounts with scores lower than $T_l$ (detected as non-fraudsters). 
For both problems a decision is taken for each account, while the proportion of wrong detections should be kept under control. Therefore the statistical performance is quantified by means of the False Discovery Rate (FDR) \cite{https://doi.org/10.1111/j.2517-6161.1995.tb02031.x} especially designed to tackle such problems.
\begin{remark}
Let us notice that FDR is related to the so-called \emph{precision} criterion in such a way that keeping FDR below a level $0<\alpha<1$ is equivalent to keeping the precision value larger than $1-\alpha$. 
However compared to other possible metrics such as the recall or F1-score, it turns out that FDR is the only one for which a data-driven procedure does exist (\cite{https://doi.org/10.1111/j.2517-6161.1995.tb02031.x}) which reaches the desired FDR control at a prescribed level $\alpha$.
\end{remark}

\paragraph{FDR performance measure}
For each account $0\leq i\leq N-1$, a score $s_i$ has been been previously computed, as described in Section~\ref{sec:representation_to}. Based on this score $s_i$, a decision is formally taken regarding respectively the threshold $T_h$ (respectively the threshold $T_l$): The account is declared as fraudster if $s_i \geq T_h$ (resp. as non-fraudster if $s_i\leq T_l$). 
In the present context (unlike the scenario described in the Section~\ref{sec.representation.learning}), the training dataset is made of labeled data $(x_i,y_i)$ where $y_i \in\acc{0,1}$ encoding the true status of the account $i$.
For detecting potential fraudsters (by computing $T_h$), a false positive is therefore an account $i$ with label $y_i=0$ and score $s_i\geq T_h$.
This leads to define the FDR value for threshold $T_h$, denoted by $FDR_h$, as
\begin{align}\label{def.FDR.h}
    FDR_h = FDR(T_h) = \mathbb{E}\croch{ \frac{\sum_{i=0}^{N-1} (1-y_i) \1_{\paren{s_i\geq T_h}}}{ \max\paren{1,\sum_{i=0}^{N-1} \1_{\paren{s_i\geq T_h}} } } },
\end{align}
where $\1_{A}$ denotes the indicator function of the event $A$. The interested reader is referred to the seminal paper \cite{https://doi.org/10.1111/j.2517-6161.1995.tb02031.x} where FDR is detailed and analyzed.
By analogy with Eq.~\eqref{def.FDR.h}, the FDR value for the $T_l$ threshold, denoted by $FDR_l$, is given by 
\begin{align}\label{def.FDR.l}
    FDR_l = FDR(T_l) = \mathbb{E}\croch{ \frac{\sum_{i=0}^{N-1} y_i \1_{\paren{s_i\leq T_l}}}{ \max\paren{1,\sum_{i=0}^{N-1} \1_{\paren{s_i\leq T_l}} } } }.
\end{align}

\paragraph{Benjamini-Hochberg (BH) procedure}
In the present work, the calibration of the thresholds $T_l$ and $T_h$ is made by using a classical multiple testing procedure called the Benjamini–Hochberg (BH) procedure (\cite{https://doi.org/10.1111/j.2517-6161.1995.tb02031.x,10.1214/aos/1013699998}).
Focusing on the high threshold $T_h$, the BH-procedure outputs a data-driven value for $T_h$ such that the resulting strategy -- detecting as fraudsters all accounts with a score larger than $T_h$ -- enjoys an FDR value  controlled at level $\alpha \in [0,1]$.

The BH-procedure relies on a $p$-value $p_i^h$ computed for an account $i$ and quantifying how much likely this account is a fraudster compared to non-fraudster accounts. Mathematically, $p_i^h$ is given by
\begin{align*}
    p_i^h = \mathbb{P}_{s^\prime \sim P_0 }\paren{ s^\prime > s_i} ,
\end{align*}
where $P_0$ denotes the probability distribution of the scores of non-fraudster accounts.
As emphasized above, such a $p$-value depends on the \emph{unknown} distribution $P_0$. But fortunately the present supervised context allows for estimating this $p$-value by
\begin{align*}
    \hat p_i^h = \frac{\sum_{j\neq i}(1-y_j) \1_{(s_j >s_i)} }{\sum_{\ell \neq i} (1-y_\ell)}, \qquad 0\leq i\leq N-1.
\end{align*}
Let us recall that $\sum_{\ell \neq i} (1-y_\ell)$ equals the number of scores among the training set data that are non-fraudsters (since $T_h$ is presently considered).
A similar definition of the estimated $p$-values can consistently be formulated for $T_l$ as follows
\begin{align*}
    \hat p_i^l = \frac{\sum_{j\neq i}y_j \1_{(s_j<s_i)} }{\sum_{\ell \neq i} y_\ell}, \qquad  0\leq i\leq N-1,
\end{align*}
where $\sum_{\ell \neq i} y_\ell$ denotes the total number of fraudsters within the training set.





Once the p-values have been estimated according to the methodology detailed above, the BH procedure is applied to compute the two required thresholds $T_h$ and respectively $T_l$.
Focusing on $T_h$ the BH-procedure starts by ranking the estimated $p$-values $\hat p^h_{(1)} \leq \cdots \leq \hat p^h_{(N)} $, where the notation $\hat p^h_{(i)}$ refers to the $i$th smallest $p$-value within the set $\acc{\hat p^h_{(i)}, \ldots, \hat p^h_{(N)}}$. Then for a prescribed level $\alpha \in (0,1)$, the number of rejections is given by
\begin{align}
    i^h_{BH,\alpha} = i^h_{BH}(\alpha) = \max\acc{ 1\leq i \leq N \mid \hat p^h_{(i)} \leq i\alpha/N }.
\end{align}
Next Figure~\ref{fig:train_test_set} illustrates how the BH-procedure works on an example. The right-most location where the straight line crosses the $p$-values curve indicates the number of rejections at level $\alpha$ ($i^h_{BH,\alpha} $), while the corresponding value on the $y$-axis denotes the value of $ \hat p^h_{(i^h_{BH,\alpha})}$. 
\begin{figure}[H]
  \centering
\includegraphics[width=0.75\textwidth]{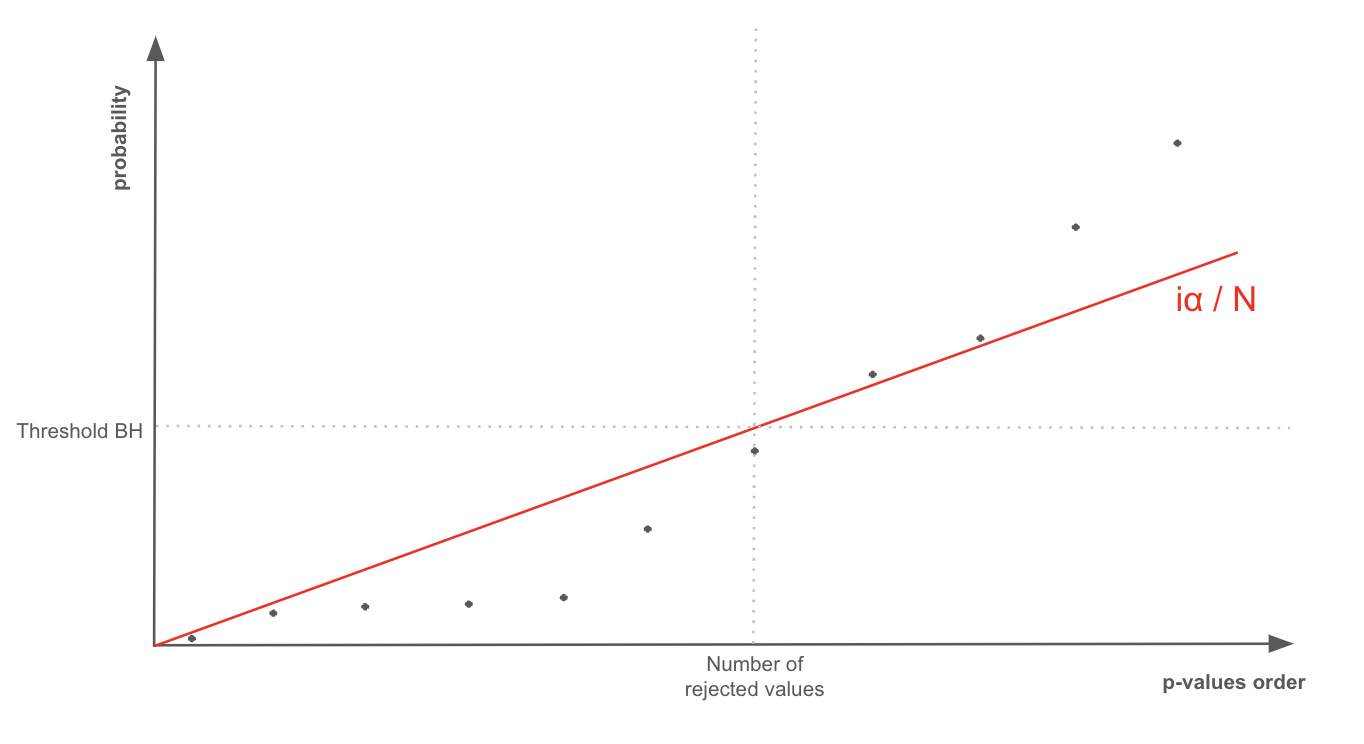}
  \caption{BH procedure and ranked p-values}
  \label{fig:threshold_bh}
\end{figure}
The threshold $T_h$ is then given by plugging the BH index $i^h_{BH,\alpha}$ in the ordered score that is,
\begin{align}\label{eq.BH.threshold.high}
    T_{h,\alpha} = T_h(\alpha) = s_{(i^h_{BH,\alpha})},
\end{align}
where $s_{(1)}\geq \dots \geq s_{(N)}$ denote the non-increasing observed scores.
Importantly \cite{https://doi.org/10.1111/j.2517-6161.1995.tb02031.x} established that, with independent $p$-values following continuous probability distributions, rejecting all the accounts with a score larger than or equal to $T_{h,\alpha}$ results in
\begin{align}
    FDR_{h,\alpha} = FDR_{T_{h,\alpha}} = \frac{\sum_{i=0}^{N-1} (1-y_i)}{N} \alpha \leq \alpha.
\end{align}
Let us emphasize that the actual FDR level is $\frac{\sum_{i=0}^{N-1} (1-y_i)}{N} \alpha$, which can be strongly lower than $\alpha$ in cases where $\frac{\sum_{i=0}^{N-1} (1-y_i)}{N}\ll1$. As long as one considers the high threshold $T_h$, the latter ratio equals the proportion of non-fraudsters among the observations. 
In the money laundering context, only a small number of fraudsters is actually expected. Therefore $\frac{\sum_{i=0}^{N-1} (1-y_i)}{N} \approx 1$ and there should not be any large difference between $\alpha$ and the actual FDR control as long as one considers $T_h$.

However the situation is completely reversed when considering $T_l$ since in this case, the actual FDR control is $\frac{\sum_{i=0}^{N-1} y_i}{N} \alpha$, where 
$\frac{\sum_{i=0}^{N-1} y_i}{N}$ denotes the proportion of fraudsters. As long as one considers the low threshold $T_l$, $\frac{\sum_{i=0}^{N-1} y_i}{N} \ll 1$ and one needs to take this phenomenon into account for avoiding a too conservative decision \cite{storey2002direct,celisse2010cross}. This can be overcome by adjusting the targeted control level $\alpha^\prime$ by choosing $\alpha^\prime = N/\sum_{i=0}^{N-1} y_i \alpha$ in the case of the low threshold.
Indeed plugging in this value of $\alpha^\prime$ within the BH-procedure leads to the desired level since
\begin{align} \label{p_value_l}
    FDR_{l,\alpha^\prime} = \frac{\sum_{i=0}^{N-1} y_i}{N} \alpha^\prime = \alpha.
\end{align}
As long as supervised observations are available, the above correction is obvious. However without supervision, there still exist approaches for estimating the proportion of fraudsters based on the known probability distribution of $p$-values under the null hypothesis as explained for instance in \cite{storey2002direct,celisse2010cross}.




\begin{remark}
    Let us emphasize that the BH-procedure relies on the last crossing-point between the straight line with slope $\alpha^\prime/N$ and the $p$-values curve. 
    For defining a threshold $T_h$ such that the corresponding FDR value is equal to the prescribed level, it is then necessary that such a crossing-point does exist. In situations where $\alpha^\prime$ is too small, then it can happen that no such crossing-point does exist. Then it results in no rejections, meaning that the desired level is so small that rejecting no ones is the only (degenerate) solution to this problem.
\end{remark}

Algorithm~\ref{algo.BH} describes in pseudo-code the successive steps corresponding to the application of the BH-procedure.
\begin{algorithm}[H]
\caption{Pseudo code of the BH-procedure for $T_h$\label{algo.BH}}
\begin{algorithmic}[2]

\STATE {\bf Input:} List of $p$-values: $\hat p^h_{0}, \ldots, \hat p^h_{N-1}$, level $\alpha \in(0,1)$, labels: $\paren{ y_0,\ldots,y_{N-1}} \in \acc{0,1}^N$. \\
\quad Compute $\alpha' = \frac{N}{N_l} \alpha$, with $N_l = \sum_{i=0}^{N-1} y_i$. \\
\quad Sort the $p$-values: $\hat p^h_{(1)}\leq \ldots \leq \hat p^h_{(N)}$.\\
\quad Define the list $pValueSorted = (\hat p^h_{(1)}, \ldots ,\hat p^h_{(N)})$. \\
\quad Define the list $CriticalValues = \paren{\alpha^\prime/N, 2\alpha^\prime/N,\ldots,\alpha^\prime}$.
\\
\quad Define $\mathcal{R}=\emptyset$.

\FOR{$i \in \croch{1, N} \cap \mathbb{N}$}
    \IF{$p^h_{(i)} \leq i \frac{\alpha^\prime}{N}$}
        \STATE $\mathcal{R} \leftarrow i$
    \ENDIF
\ENDFOR

\quad Compute $i_{BH, \alpha}^h = max(\mathcal{R})$. \\

{\bf Output:} $i_{BH, \alpha}^h$
\end{algorithmic}
\end{algorithm}


\section{Simulation experiments}
\label{sec.experiments}

The present section is devoted to the empirical assessment of the approach detailed along this work. 

Section~\ref{sec.Data.Description} describes all the features of the real datasets used to make the assessment. 
Then the ability of our approach for designing a valuable representation in a latent space is studied. In the context of money laundering, a relevant representation of the accounts (in the latent space) should allow for easily distinguishing between fraudsters and non-fraudsters.
Therefore Section~\ref{sec.representation.learning.experiments} first quantifies this performance in terms of the quality of the representations' visualization, at the latent space level. 
In a second step, Section~\ref{sec.distribution.FNF.experiments} rather focuses the classification problem where accounts are to be classified as fraudsters or non-fraudsters by means of a (basic) logistic regression classifier. 
Lastly Section~\ref{sec.learning.classif.threshold.experiments} illustrates the improvement achieved by our procedure in terms of power (number of detections): For a given FDR level, our approach leads to more detections compared to other competitors.

\subsection{Real-life dataset description}
\label{sec.Data.Description}

\paragraph{Complex (real-life) time series}
For the purpose of the present assessment, a dataset $\mathcal{D}= \{ x_1, x_1, x_2, \ldots, x_{N} \}$ has been created based on real-life data (see also Appendix~\ref{appendix.dataset.details}). Each observation $x_i = (x_{i,1},\dots,x_{i,T})$ is a bank account described by a time series of length 3 months collecting all financial transactions $x_{i,t}$ along this period. It is worth emphasizing that the resulting time series of each account is highly complex. Indeed at each time where a transaction occurs, the latter is described by both quantitative and qualitative descriptors, making the encoding of such an object highly challenging. 
More precisely every transaction is described by both quantitative and qualitative descriptors such as:
\begin{itemize}
    \item Value associated with the transaction (quantitative feature).
    \item Direction of the transaction (payin or payout, qualitative feature)
    \item Type of payment (a card payment or a transfer for example, qualitative feature)
    \item Country associated with the counterparty (qualitative feature)
    \item Some descriptors indicating if there are some keywords explaining the transaction (in a description field for example, qualitative feature)
    \item \dots
\end{itemize}
In the present context, money laundering detection is made at the account/customer level that is, an account is declared or not as fraudster based on a set of suspicious transactions with the attributes described above along a given period of time. In particular on the one hand, the present procedure does not aim at detecting specific transactions as fraud events, which can be the topic of a future work (but is out of the scope of the present one). On the other hand, it seems important to keep in mind that a given account can exhibit a fraudulent behavior on a given period of time, but can have a normal behavior outside of this period.
\begin{remark}
    Let us finally emphasize that to the best of our knowledge, such a real-life dataset does not already exist in the literature. The dataset has been made freely available on Kaggle dataset: \url{https://www.kaggle.com/datasets/haroldg/time-series-of-transactions-money-laundering/data}
\end{remark}

\paragraph{Confidentiality}
In order to protect the confidentiality of users, the data have been anonymized using differential privacy techniques combined with a PCA transformation \cite{Imtiaz2016SymmetricMP, diffprivlib}. For this procedure, the privacy parameter $\epsilon$ is equal to 1.
After such an anonymization process, the data only reflect the generic financial activity of a set of accounts corresponding to companies from various industrial domains over a period of 3 months. 

\paragraph{Train/test sets}
The dataset has been split into two subsets corresponding to a train and a test set.
Appendix~\ref{appendix.dataset.details} describes with more details the construction of the train and the test sets from the dataset $\mathcal{D}$.
Assuming that the relevant descriptors of normal time series (as opposed to time series of abnormal accounts) should not vary along the time, the training and test sets have been created by considering two disjoint time periods.
The model is trained at a given reference date based on the elapsed period until that date. The predictions are made on time series of 3 months of transactions, from a period of 6 months following the reference date as illustrated by Figure~\ref{fig:train_test_set} below.
\begin{figure}[H]
  \centering
  \includegraphics[width=0.75\textwidth]{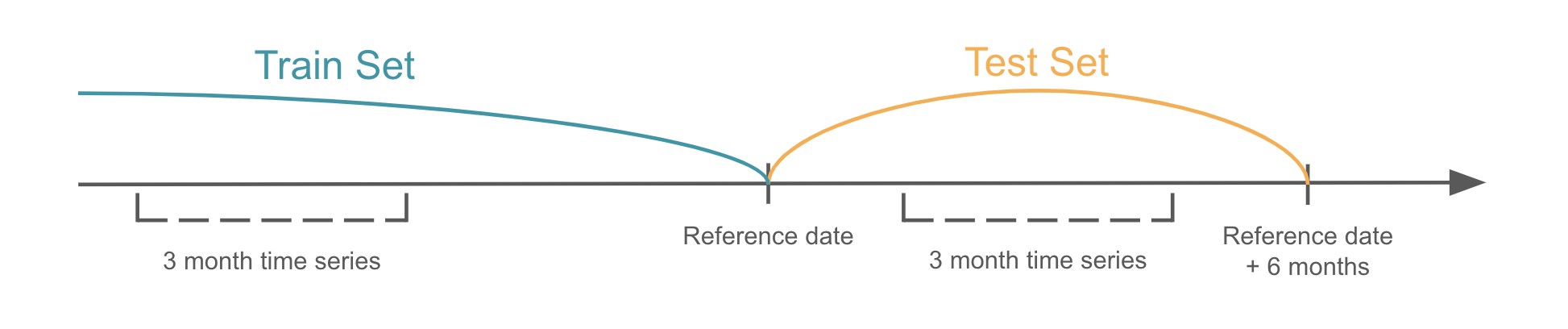}
  \caption{Train set and test set}
  \label{fig:train_test_set}
\end{figure}

\paragraph{Strongly imbalanced data}
Here the goal is to detect accounts (companies) involved in money laundering. One important specificity of our dataset is its strong imbalance regarding the (expected) very low proportion of fraudsters among all observations. For instance the training set contains 27\% of fraudsters (which is not so bad), while the test set only contains 5\% of fraudulent companies (which is clearly worse). 
Let us recall that each account is labeled as non-fraudster or fraudster based on a running detection process during each considered period. An account can be labeled as non-fraudster in the training set and can be flagged as fraudster in the test set. In other words the occurrence of fraudulent events vary along the time, which explains the variation of the proportion of fraudsters in the test set compared to the training set.


Let us also mention that due to confidentiality requirements, the cardinality of the designed dataset is small compared to all the data available. 
Another specificity of this dataset owes to the sampling rate of the fraudsters. Actually the proportion of fraudsters is higher in the current dataset than the one of the full dataset. We recall that the current dataset is only a small fraction of the full real-life dataset. Detecting fraudsters would be almost impossible if the fraudsters proportion is too low as well (at least as low as in the full dataset). The main reason for increasing the proportion of fraudsters in the current dataset is to help the model comparison. Obviously if the proportion of fraudsters is so low that all models perform equally poorly, drawing any meaningful conclusion would be almost impossible.

\paragraph{Noisy labels}
As mentioned before, the datasets are real-life ones. The current labeling protocols define the "non-fraudulent" observations as observations that are not yet detected. But, it is possible to have "non-fraudulent" observations that are "fraudulent" in reality (until a potential future detection). This ambiguity necessitates detection frameworks robust to label noise (possible label mistakes).

\paragraph{Using aggregates}

From the full dataset, another dataset denoted by $\mathcal{D}^\prime$ is designed as well. It contains the same number of observations/accounts as $\mathcal{D}$, but each account is now described by more classical aggregates globally reflecting the full sequence of transactions along the considered time period 
\cite{Tertychnyi2020ScalableAI}.
Instances of these aggregates are the summation, the maximum or the minimum of the payouts along the past months to name but a few.
It should be clear that such aggregates describe less precisely the account behaviors than the time series used in $\mathcal{D}$. One reason is that they do not incorporate any information regarding the \emph{sequential structure} of transaction events, which can be highly informative in practice for money laundering detection. 

Also in addition to these aggregates, each account $x^\prime_i$ in $\mathcal{D}^\prime$ has some high-level descriptors of the customer such its legal form, its activity category (NAF code in France for exemple).
%


\subsection{Representation learning visualization}
\label{sec.representation.learning.experiments}

\begin{figure}[H]
  \centering
  \includegraphics[width=0.6\textwidth]{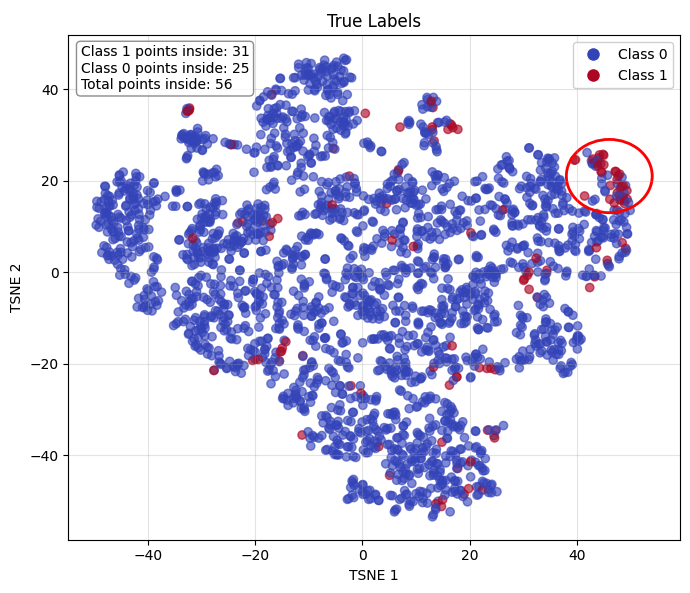}
  \caption{Embeddings of the transformer projected with t-SNE - Comparison between money laundering labels (class 0: Non-Fraudster and class 1: Fraudsters)}
  \label{fig:transformer-tsne}
\end{figure}

From a candidate representation -- called an embedding -- of all observations in the latent space, the true labels can be used for measuring the "homogeneity" of the classes once displayed at the latent space level.
Figure~\ref{fig:transformer-tsne} illustrates the transformer's embedding locations in a (projected) latent space, revealing a cluster of fraudsters in the top-right corner (within the red circle). Multiple training runs with different random seeds systematically highlight one or two such compact fraudster clusters over small regions in the latent space. This visualization represents a typical outcome from contrastive pre-training, showing the transformer's ability to capture specific features among the descriptors that are related to suspicious patterns.

\begin{figure}[H]
  \centering
  \includegraphics[width=0.6\textwidth]{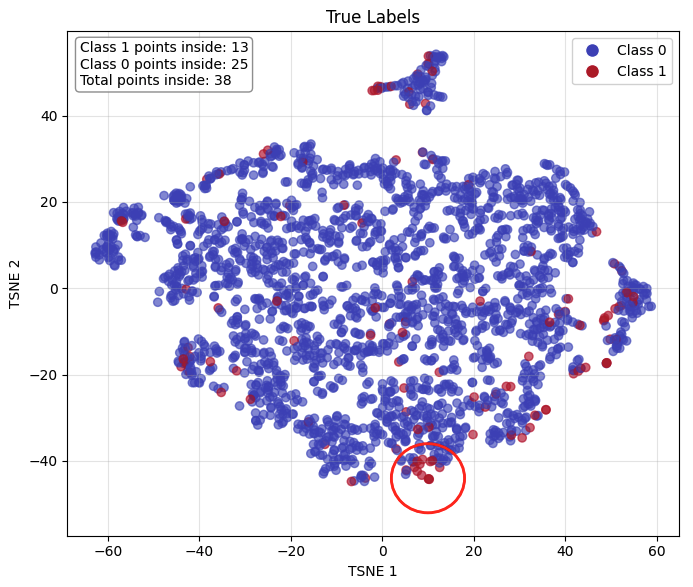}
  \caption{Embeddings of the LSTM Autoencoder projected with t-SNE - Comparison between money laundering labels (class 0: Non-Fraudster and class 1: Fraudsters)}
  \label{fig:lstm_autoencoder_tsne}
\end{figure}

In contrast, Figure~\ref{fig:lstm_autoencoder_tsne} displays representations output by the LSTM Autoencoder \cite{srivastava2016unsupervisedlearningvideorepresentations} (see Section~\ref{sec:related.works.in.representation.learning}). 
Whereas it is possible to identify a cluster of fraudsters at the bottom of the picture, fewer fraudsters belong to this region compared to Figure~\ref{fig:transformer-tsne}. 
In addition, representations exhibit a greater variability across the successive runs (for different seeds) compared to the transformer one.

Appendix~\ref{appendix.representation.visualization} provides a set of visualizations across different random seeds, highlighting the transformer improvement in terms of stability of the results.
This is consistent with the results presented in Sections~\ref{sec.distribution.FNF.experiments} and~\ref{sec.learning.classif.threshold.experiments}, where the transformer achieves higher detection rates.

\subsection{Distributions of fraudsters and non-fraudsters}
\label{sec.distribution.FNF.experiments}

One of the main goals of the present work is to distinguish the sub-populations of fraudsters and non-fraudsters from the initial observations.
After a representation (equivalently an embedding) has been learned in a latent space from Section~\ref{sec:representation_to}, a classification rule (logistic regression) is fed up with this representation for classifying the observations as fraudsters or non-fraudsters.
The following reviews the performance of the embeddings output by the LSTM Autoencoder \cite{srivastava2016unsupervisedlearningvideorepresentations} and the transformer combined with the logistic regression classifier \cite{https://doi.org/10.1111/j.1467-9574.1988.tb01237.x, Tertychnyi2020ScalableAI} used for answering the money laundering detection task.  
The tabular dataset $\mathcal{D^\prime}$ is also considered as an input for the logistic regression and an XGBoost \cite{Chen_2016}, allowing for a comparison with domain-informed features.
To go further, as described along Section~\ref{sec:from.representations.to.aml}, the transformer is also fine-tuned on the labels.

\begin{figure}[H]
  \centering
  \includegraphics[width=1\textwidth]{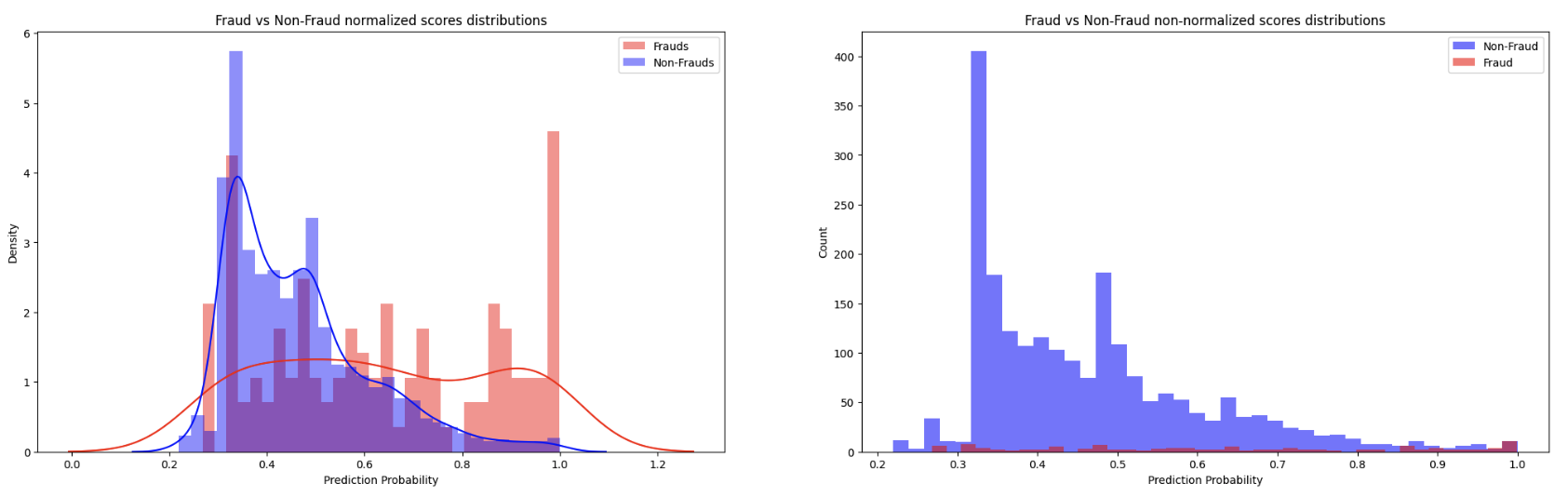}
  \caption{Normalized and non-normalized classification score distribution - Logistic Regression on tabular data only}
  \label{fig:tab_approach_logistic_score_distribution}
\end{figure}

A specific difficulty raised by the money laundering detection is the strong imbalance between classes: fraudsters are at most 27\% of the observations in the training set, and only 5\% in the test set (see Section~\ref{sec.Data.Description}).
This leads us to replace the usual classification metrics by rather assessing how much the scores output by the logistic regression exhibit a different behavior depending on the true labels.
A first illustration of the two distributions output by logistic regression is given by Figure~\ref{fig:tab_approach_logistic_score_distribution} (Left panel), where only tabular data have been used to learn the two distributions. 
Two histograms are displayed: red for the distribution of (the scores of) fraudsters, and blue for non-fraudsters. The continuous curves on the picture are the kernel density estimators respectively of the fraudsters distribution (red) and non-fraudsters distribution (blue) learned with a Gaussian kernel.
An effective approach should lead to a probability distribution of fraudsters (red) concentrated on the right (corresponding to high scores), whereas the distribution of non-fraudsters (blue) should be shifted to the left. Intuitively the smaller the overlap between the two distributions, the more effective the underlying methodology (cf. Figure~\ref{fig:Score.distribution}).

\begin{figure}[H]
  \centering
  \includegraphics[width=1\textwidth]{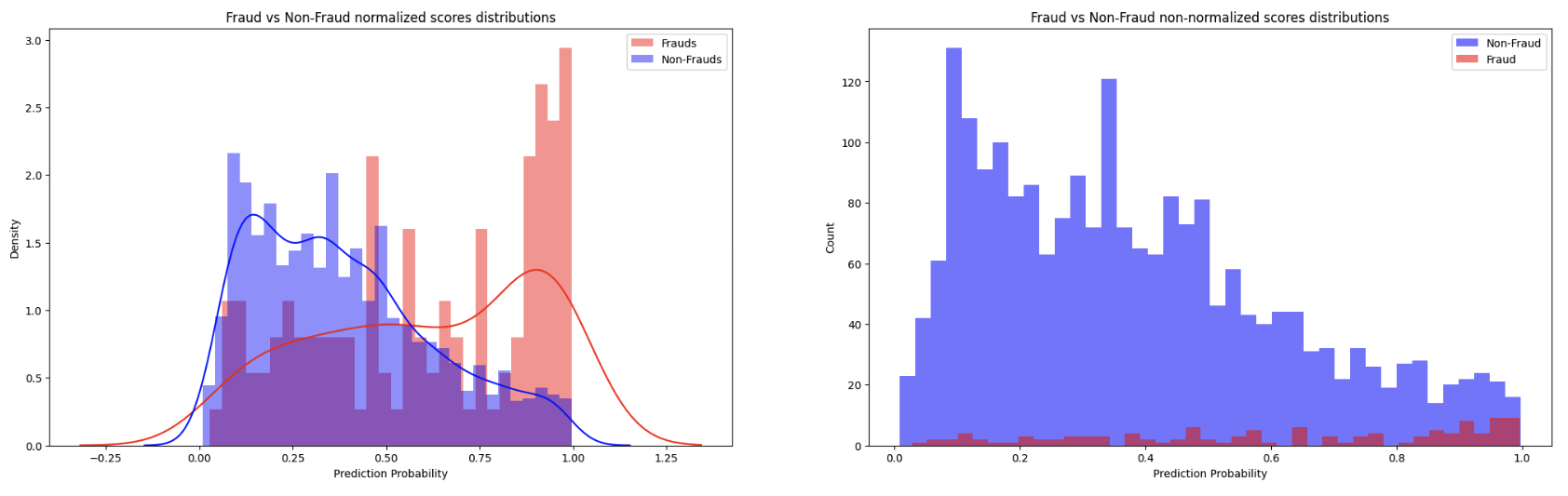}
  \caption{Normalized and non-normalized classification score distribution - XGBoost on tabular data only}
  \label{fig:tab_approach_xgboost_distribution}
\end{figure}

Figures~\ref{fig:tab_approach_logistic_score_distribution} and \ref{fig:tab_approach_xgboost_distribution} demonstrate the performance of logistic regression and of XGBoost on the tabular dataset $\mathcal{D^\prime}$. 
From the left panel of Figure~\ref{fig:tab_approach_logistic_score_distribution}, it appears that tabular data from $\mathcal{D}^\prime$ only lead to a poor embedding. Indeed the non-fraudsters distribution truly exhibits a bump on the left. However the fraudsters distribution does not clearly exhibit any similar bump on the right. Moreover the supports of the two probability distribution are almost coinciding. As a consequence any detection rule which would consist in detecting as fraudsters all observations with a score larger than a threshold would unavoidably lead to a large number of false positives. 
This conclusion is also supported by the right panel of Figure~\ref{fig:tab_approach_logistic_score_distribution} where the counts of each class are displayed. On the right tail (in $[0.85;1]$) of the bar plot, non-fraudsters and fraudsters are mixed with an almost equal cardinality. 
The weak ability of the procedure for distinguishing the two populations actually reflects the poor quality of the aggregates from $\mathcal{D}^\prime$ (see Section~\ref{sec.Data.Description}) that have lead to these embeddings. Therefore such aggregates are not informative enough to summarize complex time series such as the ones considered in $\mathcal{D}$. Nevertheless such a rough representation could be seen as a first improvement compared to ongoing rule-based strategies for raising alerts in money laundering detection.

A similar analysis can be performed on the Figure~\ref{fig:tab_approach_xgboost_distribution}, displaying the results of an XGBoost classifier on the dataset $\mathcal{D}^\prime$.
From the left panel, it appears that the model is able to exhibit two bumps for the fraudster and non-fraudster distributions. Especially, the bump associated to the fraudsters on the left is more pronounced than the one seen with logistic regression in Figure~\ref{fig:tab_approach_logistic_score_distribution}.
But on the right panel, the results are worse, leading to a higher false positive rate (which is confirmed in the results later displayed in Section~\ref{sec.learning.classif.threshold.experiments}).
Again, the supports of the two probability distributions are almost coinciding (similarly to the logistic regression, but even more). As a consequence, any detection rule which would consist in detecting as fraudsters all observations with a score larger than a threshold would unavoidably lead to a large number of false positives.

\begin{figure}[H]
  \centering
  \includegraphics[width=1\textwidth]{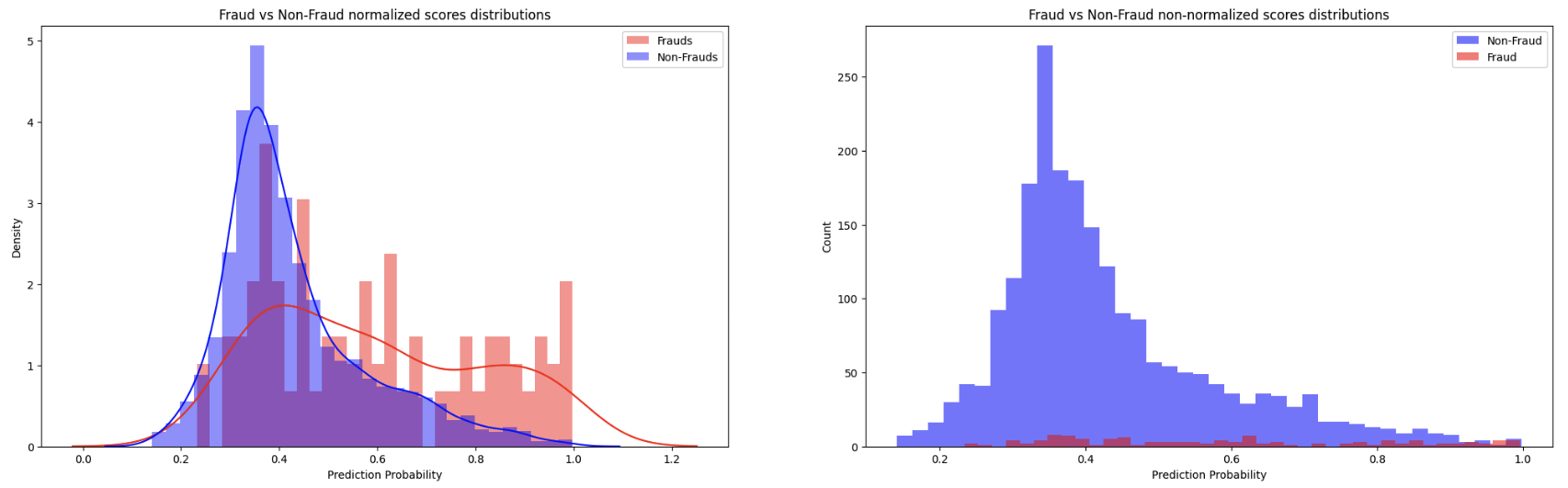}
  \caption{Normalized and non-normalized classification score distribution - Logistic Regression on LSTM autoencoder representations}
  \label{fig:lstm_autoencoder_logistic_score_distribution}
\end{figure}
By contrast, Figure~\ref{fig:lstm_autoencoder_logistic_score_distribution} displays the results obtained from an LSTM autoencoder \cite{srivastava2016unsupervisedlearningvideorepresentations}, where the complex time series from $\mathcal{D}$ are used instead of their proxies in $\mathcal{D}^\prime$. The LSTM autoencoder parameters have been learned from $\mathcal{D}$ and hyper-parameters have been tuned with a grid search strategy.
By comparison between the left panels of Figures~\ref{fig:tab_approach_logistic_score_distribution},~\ref{fig:tab_approach_xgboost_distribution}, and~\ref{fig:lstm_autoencoder_logistic_score_distribution}, one observes similar (and still relatively poor) results for the LSTM autoencoder, demonstrating the complexity of handling money laundering detection based on the complex dataset $\mathcal{D}$.
Structured time series are more informative regarding the problem at hand. However, the model is not able to leverage efficiently the time series to extract money laundering patterns.
The LSTM results are included in the comparison to serve as a baseline when introducing our approach based on a transformer learned by contrastive learning and applied to the dataset $\mathcal{D}$.

\begin{figure}[H]
  \centering
  \includegraphics[width=1\textwidth]{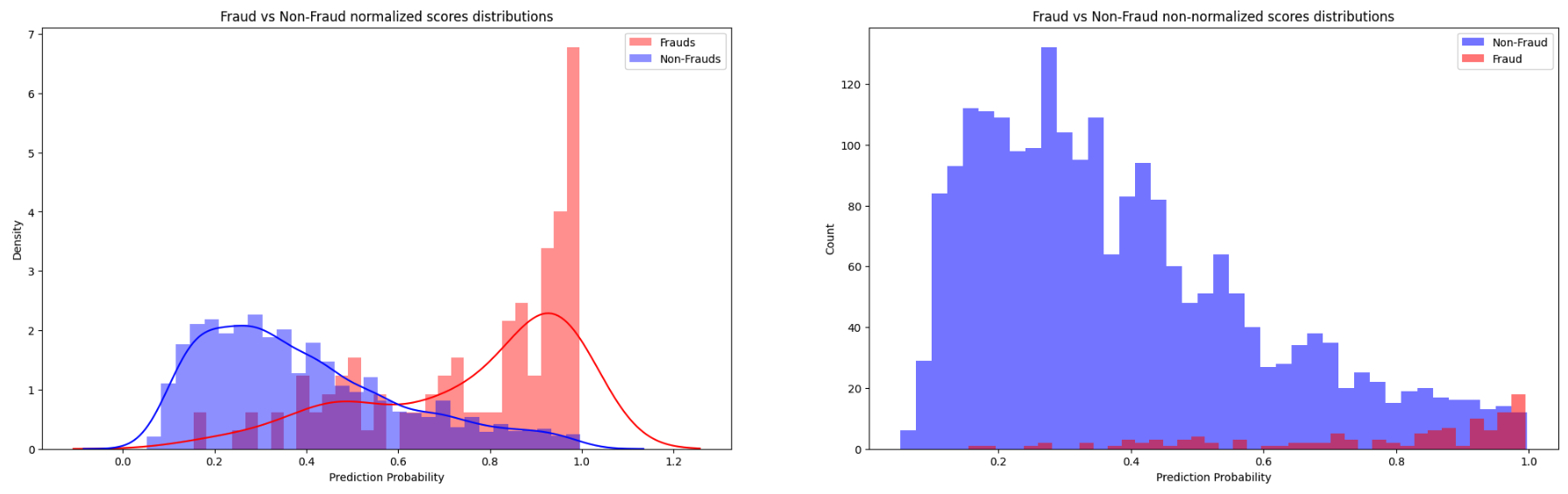}
  \caption{Normalized and non-normalized classification score distribution - Logistic Regression on contrastive learned representations}
\label{fig:contrastive_representations_logistic_score_distribution}
\end{figure}

The performance of the transformer-based approach are displayed in Figure~\ref{fig:contrastive_representations_logistic_score_distribution}.
The transformer is optimized for learning an embedding of the observations in $\mathcal{D}$. This learning step is made using contrastive learning and data in $\mathcal{D}^\prime$ since no label from $\mathcal{D}$ is used at the representation learning step (See Section~\ref{sec.representation.learning} and Section~\ref{sec:representation_to} for more details). 
The hyperparameters from Eq.~\ref{eq_representations_as_features} such as the learning rate and temperature are calibrated using a grid search strategy. 

The left panel of Figure~\ref{fig:contrastive_representations_logistic_score_distribution} shows two main improvements compared to the previous graphs. 
In the left tail of the non-fraudsters distribution (blue), the width of the non-overlapping part is visually much bigger than the previous approaches, with no overlapping before a score of $0.4$ and a lot of observations in $[0.0;0.4]$.
This is also visible from the corresponding right panel of Figure~\ref{fig:contrastive_representations_logistic_score_distribution} where no fraudsters lie in this range.
The second visible improvement is the higher concentration of fraudsters in a close neighborhood to 1 (higher bump), with a very low density for non-fraudsters in this area. Combining these two features makes the two classes more distinguishable. It therefore confirms that the transformer trained with contrastive learning is more powerful than the LSTM autoencoder for designing an effective embedding of the complex financial time series in the context of money laundering detection.

For the sake of completeness, the fine tuning of the transformer has been also included in the comparison. 
It only differs from the previous approach by a new optimization step of the transformer parameters when the latter is combined with a logistic regression classifier. The former parameters learned from the previous step now serve as an (informative) initialization of the parameters (for hopefully avoiding bad local minima). The optimization procedure has now been adapted for the classification task.
Here the purpose is to check whether there is (or not) something to gain in re-optimizing the transformer parameters compared to the results previously discussed from Figure~\ref{fig:contrastive_representations_logistic_score_distribution}.
\begin{figure}[H]
  \centering
  \includegraphics[width=1\textwidth]{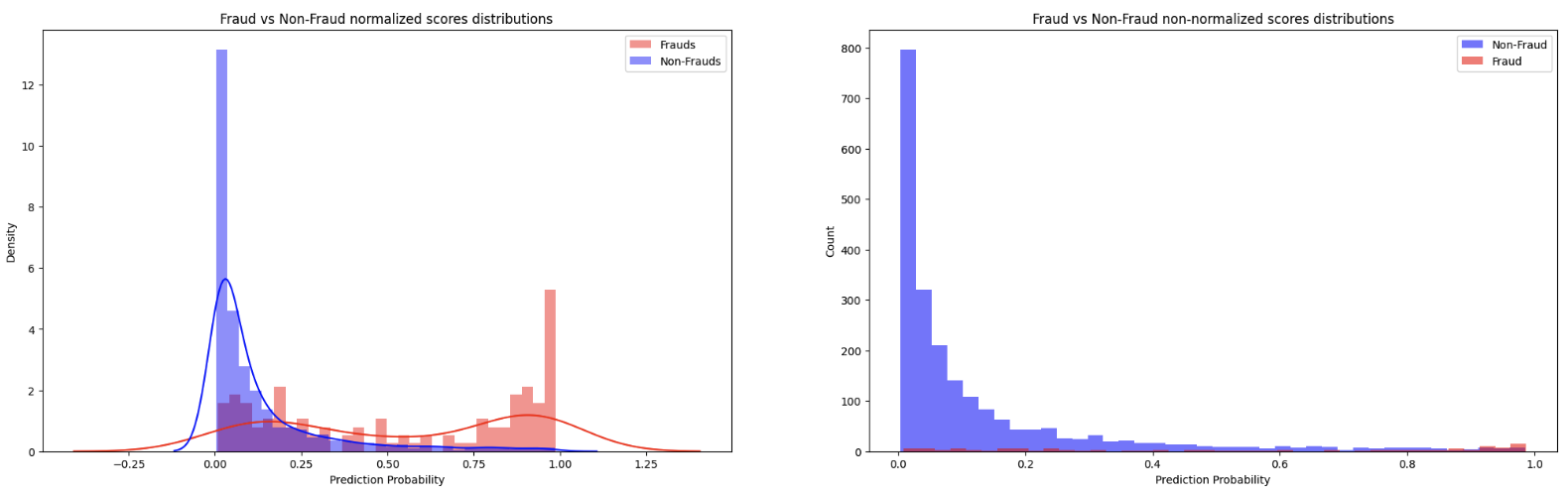}
  \caption{Normalized and non-normalized classification score distribution - Transformer fine-tuned after contrastive pre-training}
  \label{fig:fine_tuned_distributions_score}
\end{figure}
From the left panel of Figure~\ref{fig:fine_tuned_distributions_score} one clearly observe a strong change regarding the non-fraudsters probability distribution shape. Compared with the one of Figure~\ref{fig:contrastive_representations_logistic_score_distribution}, it is now more peaky in a close neighborhood of 0. Moving to the right panel, it appears that only a small fraction of the non-fraudsters population has a score above 0.4, which deeply contrasts with the right panel of Figure~\ref{fig:contrastive_representations_logistic_score_distribution}. 
However the improvement for the fraudsters distribution is unfortunately not as clear.
This may result from the strong imbalance in our data between the class of non-fraudsters (at least 75\%) and the one of fraudsters (at most 25\% in the training set).

\subsection{Calibrating classification thresholds}
\label{sec.learning.classif.threshold.experiments}

Illustrations such as
Figure~\ref{fig:contrastive_representations_logistic_score_distribution} from previous Section~\ref{sec.distribution.FNF.experiments} highlight that defining a unique threshold for distinguishing between fraudsters and non-fraudsters would necessary lead to an overly large number of mistakes in both classes. This results from the overlapping supports of these two probability distributions.
As a consequence, Section~\ref{sec.Two.thresholds} describes a detection procedure relying on two thresholds respectively denoted by $T_h$ and $T_l$ 
respectively for high-threshold (in the right tail) and the low-threshold (left tail).
These thresholds are calibrated by means of the BH-procedure consistently with the material exposed along Section~\ref{sec.thresholds.BH}.

Our present purpose is to illustrate how this BH-procedure actually behaves in practice in terms of the number of output detections (as fraudsters for $T_h$ or as non-fraudsters for $T_l$).
This illustration is carried out with the four strategies reviewed along previous Section~\ref{sec.distribution.FNF.experiments} that is, Logistic regression with Tabular Data (LTD), XGBoost with Tabular Data (XGBTD), LSTM and logistic regression (LSTM), Transformer and logistic regression (Transf), and Fine Tuning (FT).
The results are reported in Table~\ref{tab:perf.summary.high.threshold} for $T_h$ and Table~\ref{tab:perf.summary.low.threshold} for $T_l$ from applying the above five strategies on the test set data (with detailed specifications in Section~\ref{sec.Data.Description}). 

In most settings, the BH-procedure achieves the expected result. 
However, there are a few exceptions where the result cannot be achieved since the required control level $\alpha$ is too low (too conservative control). In such a case the results are reported as "\textit{NA}" in Tables~\ref{tab:perf.summary.high.threshold} and~\ref{tab:perf.summary.low.threshold}.

\paragraph{Behavior of the High-threshold $T_h$}
In Table~\ref{tab:perf.summary.high.threshold}, the column entitled $FDR_h$ displays the required FDR level. 
Results are provided for four values of $FDR_h$, namely from $0.7$ to $0.1$. For each FDR value, the corresponding number of fraudsters is much larger for Transf and FT than for other competitors. Depending on $FDR_h$, it is twice larger than the number of detected fraudster of LSTM (and even more for lower $FDR_h$ values). For a prescribed proportion of false detections, Transf and FT uniformly identify more fraudsters than XGBTD or TD.
%
%
\begin{table}[H]
\begin{tabular}{lllll}
Model                                  & $FDR_h$ & Frauds detected & \% total frauds & F1-score \\
\hline
Logistic Regression - tabular data & 0.40 & 0.0 $\pm$ 0.0 & 0.0 & 0.0 \\
XGBoost - tabular data    & 0.40 & 1.0 $\pm$ 0.0 & 0.86 & 0.02 \\
Logistic Regression - LSTM & 0.40 & 8.5 $\pm$ 6.2 & 7.33 & 0.13 \\
Logistic Regression - transformer & 0.40 & 23.6 $\pm$ 12.8 & 20.34 & 0.30 \\
\textbf{Transformer fine-tuning} & \textbf{0.40} & \textbf{26.7 $\pm$ 14.5} & \textbf{23.02} & \textbf{0.33} \\
\hline
Logistic Regression - tabular data & 0.30 & 0.0 $\pm$ 0.0 & 0.0 & 0.0 \\
XGBoost - tabular data    & 0.30 &0.0 $\pm$ 0.0 & 0.0 & 0.0 \\
Logistic Regression - LSTM & 0.30 & 5.5 $\pm$ 4.0 & 4.74 & 0.089 \\
\textbf{Logistic Regression - transformer} & \textbf{0.30} & \textbf{12.3 $\pm$ 11.6} & \textbf{10.60} & \textbf{0.18} \\
Transformer fine-tuning & 0.30 & 10.2 $\pm$ 9.2 & 8.79 & 0.15 \\
\hline
Logistic Regression - tabular data & 0.20 & 0.0 $\pm$ 0.0 & 0.0 & 0.0 \\
XGBoost - tabular data    & 0.20 & 0.0 $\pm$ 0.0 & 0.0 & 0.0 \\
Logistic Regression - LSTM & 0.20 & 2.6 $\pm$ 1.9 & 2.41 & 0.04 \\
\textbf{Logistic Regression - transformer} & \textbf{0.20} & \textbf{4.0 $\pm$ 5.0} & \textbf{3.45} & \textbf{0.06} \\
Transformer fine-tuning & 0.20 & 0.8 $\pm$ 2.5 & 0.69 & 0.01 \\
\hline
Logistic Regression - tabular data & 0.10 & 0.0 $\pm$ 0.0 & 0.0 & 0.0 \\
XGBoost - tabular data    & 0.10 & 0.0 $\pm$ 0.0 & 0.0 & 0.0 \\
Logistic Regression - LSTM & 0.10 & 0.0 $\pm$ 0.0 & 0.0 & 0.0 \\
\textbf{Logistic Regression - transformer} & \textbf{0.10} & \textbf{0.9 $\pm$ 2.8} & \textbf{0.78} & \textbf{0.01} \\
Transformer fine-tuning & 0.10 & 0.0 $\pm$ 0.0 & 0.0 & 0.0 \\
\hline
\end{tabular}
\caption{Performance summary on the high threshold}
\label{tab:perf.summary.high.threshold}
\end{table}
Let us emphasize that the large $FDR_h$ values considered in Table~\ref{tab:perf.summary.high.threshold} reflect the fact that the distributions of fraudsters and non-fraudsters are strongly overlapping in the right tail of Figure~\ref{fig:contrastive_representations_logistic_score_distribution} and fraudsters are only a small fraction of the full observations (5\% in the test set). Since the supports are not disjoint close to 1, detecting each true fraudster can only be made at the price of including quite a few non-fraudsters. 
Nevertheless it is important to keep in mind that this is already by far better than the state-of-the-art rule-based approaches currently used to answer this question.

\paragraph{Behavior of the Low-threshold $T_l$}
Results for $T_l$ are collected in Table~\ref{tab:perf.summary.low.threshold}. 
By contrast with the (large) FDR values from Table~\ref{tab:perf.summary.high.threshold}, the four values considered here $0$, $1\%$, $2\%$, and $3\%$ are usual ones. 
Let us also notice that the largest possible value of $FDR_l$ is smaller or equal to 5\%, which results from the strong imbalance between classes: Rejecting all observations as non-fraudsters would result in at most 5\% of false positives (that is, the proportion of fraudsters among the observations in the test set). 
\begin{table}[H]
\begin{tabular}{lllll}
Model & $FDR_l$ & Non-frauds detected & \% non-frauds & F1-score \\
\hline
Logistic Regression - tabular data & 0.03 & 1807.3 $\pm$ 3.6 & 82.56 & 0.89 \\
XGBoost - tabular data & 0.03 & 1766.0 $\pm$ 0.0 & 80.68 & 0.88 \\
Logistic Regression - LSTM & 0.03 & 1858.0 $\pm$ 138.0 & 84.88 & 0.91 \\
Logistic Regression - transformer & 0.03 & 2118.3 $\pm$ 11.0 & 96.77 & 0.97 \\
\textbf{Transformer fine-tuning} & \textbf{0.03} & \textbf{2125.8 $\pm$ 10.5} & \textbf{97.11} & \textbf{0.97} \\
\hline
Logistic Regression - tabular data & 0.02 & NA & NA & NA \\
XGBoost - tabular data & 0.02 & 104.0 $\pm$ 0.0 & 4.8 & 0.09 \\
Logistic Regression - LSTM & 0.02 & 871.1 $\pm$ 476.4 & 39.79 & 0.57 \\
\textbf{Logistic Regression - transformer} & \textbf{0.02} & \textbf{1932.8 $\pm$ 35.9} & \textbf{88.30} & \textbf{0.93} \\
Transformer fine-tuning & 0.02 & 1903.8 $\pm$ 64.4 & 86.97 & 0.92 \\
\hline
Logistic Regression - tabular data & 0.01 & NA & NA & NA \\
XGBoost - tabular data & 0.01 & NA & NA & NA \\
Logistic Regression - LSTM & 0.01 & 74.0 $\pm$ 88.5 & 3.38 & 0.07 \\
Logistic Regression - transformer & 0.01 & 1188.3 $\pm$ 208.4 & 54.29 & 0.70 \\
\textbf{Transformer fine-tuning} & \textbf{0.01} & \textbf{1292.5 $\pm$ 144.3} & \textbf{59.05} & \textbf{0.74} \\
\hline
Logistic Regression - tabular data & 0.0 & 17.0 $\pm$ 0.0 & 0.02 & 0.015 \\
XGBoost - tabular data & 0.0 & 19.0 $\pm$ 0.0 & 0.87 & 0.02 \\
Logistic Regression - LSTM & 0.0 & 35.1 $\pm$ 30.8 & 1.60 & 0.03 \\
Logistic Regression - transformer & 0.0 & 237.5 $\pm$ 144.2 & 10.85 & 0.20 \\
\textbf{Transformer fine-tuning} & \textbf{0.0} & \textbf{419.4 $\pm$ 149.5} & \textbf{19.16} & \textbf{0.32} \\
\hline
\end{tabular}
\caption{Performance summary on the low threshold}
\label{tab:perf.summary.low.threshold}
\end{table}
Another important feature of the results is that, for a given FDR level, the transformer uniformly achieves the best performance (in terms of the number of detected true non-fraudsters). This does not only illustrate the gain in considering the Transformer instead of an LSTM autoencoder, but it also reflects that detecting non-fraudsters is easier than detecting fraudsters.
Nevertheless let us emphasize that detecting as non-fraudsters some accounts allows for saving time and computational resources in practice since it avoids considering a large number of false alarms with a very small probability of mistakes.


\section{Conclusion}

This research addresses the critical challenge of money laundering detection in financial system with a two-steps methodology, combining a transformer that processes transactions time series and an adaptive strategy based on two thresholds that control the false positive rate.
This strongly contrasts with traditional rule-based systems, which allow for auditability but definitely struggle to maintain a reasonable false positive rate and cannot adapt to evolving patterns.

Our methodology mitigates these limitations through contrastive learning. Encapsulated in a pre-training step, it enables learning representations of financial time series, reducing the dependency on available labels.
This self-supervised step captures the relevant behaviors in the time series, enabling transferability to a downstream task.
The second step, with the two-thresholds approach optimizes the investigation time: It rejects most risky observations while simultaneously discarding those that are considered as not risky. The purpose of this is to reduce the number of unnecessary investigations while keeping the false discovery rate under control.
Experimental results on real-life data demonstrate the potential enhancement allowed by representation learning in this context. It raises opportunities to outperform traditional rule-based systems.

The proposed framework exhibits some sensitivity to the hyper-parameters tuning (temperature, learning rate, noise intensity, etc.).
This requires an extensive calibration step of these hyper-parameters to find an optimal combination, which can reveal time consuming. However this calibration step has to be carried out once and can further be updated from time to time for adapting to an evolving context.
Moreover, the empirical experiments have been only carried out from a subset of the full available dataset. Despite this limitation, the reported empirical results demonstrate a consistent behavior, reducing a lot the number of false positives compared to rule-based strategies. All of this even suggests better results on larger datasets.


\pagebreak
\bibliographystyle{alpha}
\bibliography{sample}

\pagebreak

\appendix
\section{Appendix}
\subsection{dataset details}
\label{appendix.dataset.details}

In the following we provide details about the datasets used along the present work.
\begin{table}[H]
\centering
\begin{tabular}{ll}
& \textbf{Value} \\
\hline
Number of transaction in $\mathcal{D}$ - train set & 2 343 143 \\
Number of transaction in $\mathcal{D}$ - test set & 150 689 \\
Number of features (after PCA) in $\mathcal{D}$ & 43 \\
Number of companies in $\mathcal{D^\prime}$ - train set & 5428 \\
Number of companies in $\mathcal{D^\prime}$ - test set & 2305 \\
Number of features (after PCA) in $\mathcal{D^\prime}$ & 6 \\
\% of fraudsters - train set & 26.76\% \\
\% of fraudsters - test set & 5.03\% \\
\end{tabular}
\caption{Datasets details}
\label{table:data.set.details}
\end{table}


\paragraph{Train/test sets}
As described in Section~\ref{sec.Data.Description}, the train and test sets are respectively created from disjoint time periods as illustrated by Figure~\ref{fig:train_test_set}.

\paragraph{Training and testing label imbalances}
The training and testing sets have different label imbalances. The fraudsters in the test set are the one detected during a short period of time (few months). On the contrary, the train set has a "stock" of fraudsters during a long period of time (several years). 
These two amounts are then compared to the total number of observations at the reference date (for the train set) and 6 months later (for the test set). These two amounts are on the same scale, while the amount of fraudsters is very different in thre train set and in the test set, explaining the difference of imbalances.

\begin{figure}[H]
  \centering
  \includegraphics[width=0.7\textwidth]{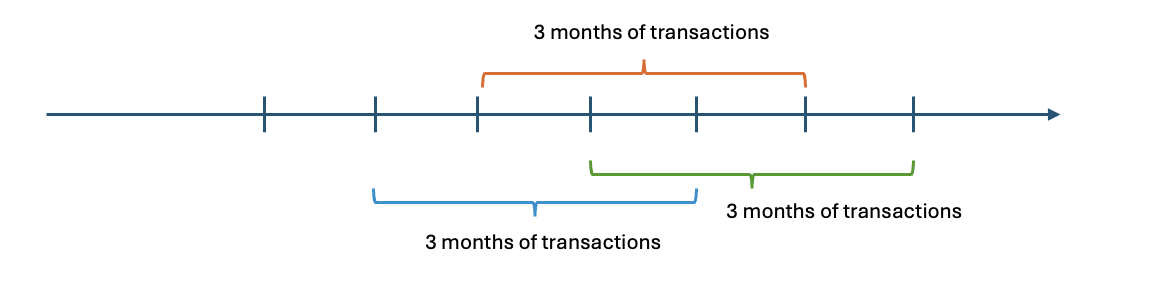}
  \caption{Transaction windows used in creating the dataset $\mathcal{D}$.}
  \label{fig:transaction.windows}
\end{figure}

\paragraph{Each firm has several time series}
Figure~\ref{fig:transaction.windows} emphasizes that every customer is associated with several time series corresponding to successive sliding windows.
%

\subsection{Hyperparameters and training details}
\label{appendix.hyperparameters.and.training.details}
We optimized the key hyperparameters — temperature, learning rate, number of layers, number of neurons, dropout rate, noise intensity, and memory‐bank size — via a grid search guided by the RankMe metric \cite{garrido2023rankmeassessingdownstreamperformance}.
These hyperparameters strongly influence the representation quality and the results reported in Section~\ref{sec.experiments}.

\paragraph{Transformer pre-training with contrastive learning}
Please find below a table describing the details about the main hyperparameters used during the contrastive pre-training of the transformer.

\begin{table}[H]
\centering
\begin{tabular}{ll}
\textbf{Hyperparameter} & \textbf{Value} \\
\hline
Projection head Dimension & 16 \\
Projection head Size & 2 \\
Memory bank size & 4000 \\
Max length & 256 \\
Number of heads & 10 \\
Number of layers & 5 \\
Number of neurons & 64 \\
Dropout rate & 0.1 \\
Number of negative examples & 128 \\
Number of positive neighbors & 50 \\
Number of epochs & 100 \\
Batch size & 512 \\
Gradient accumulation step & 8 \\
Learning rate & 0.0005 \\
Temperature & 0.2 \\
Weight decay & 0.01 \\
Optimizer & AdamW \\
Noise intensity & 0.05 \\
Gradient clipping max norm & 5.0 \\
Output representation dimension & 64 \\
\end{tabular}
\caption{Transformer hyperparmeters - Contrastive pre-training}
\label{table:hyperparameters}
\end{table}

\paragraph{Transformer fine-tuning}
Please find below a table describing the details about the main hyperparameters used during the fine tuning of the transformer, after the contrastive pre-training.
\begin{table}[H]
\centering
\begin{tabular}{ll}
\textbf{Hyperparameter} & \textbf{Value} \\
\hline
Number of epochs & 10 \\
Batch size & 1024 \\
Learning rate & 0.001 \\
Gradient clipping max norm & 5.0 \\
Optimizer & AdamW \\
\end{tabular}
\caption{Fine-tuning hyperparmeters - Contrastive pre-training}
\label{table:fine.tuning.hyperparameters}
\end{table}

\paragraph{LSTM autoencoder}
Please find below the main hyperparameters of the LSTM autoencoder, used as a benchmark in the experiments.
\begin{table}[H]
\centering
\begin{tabular}{ll}
\textbf{Hyperparameter} & \textbf{Value} \\
\hline
Number of layers & 3 \\
Hidden size & 128 \\
Embedding size & 64 \\
Dropout rate & 0.25 \\
Max length & 256 \\
Batch size & 512 \\
Optimizer & AdamW \\
Learning rate & 0.001 \\
Weight decay & 0.01 \\
Number of epochs & 100
\end{tabular}
\caption{LSTM hyperparmeters - Contrastive pre-training}
\label{table:lstm.hyperparameters}
\end{table}

\subsection{Representation learning visualization with different seeds}
\label{appendix.representation.visualization}

\subsubsection{Transformer with contrastive learning}
\begin{figure}[H]
  \centering
  \includegraphics[width=0.6\textwidth]{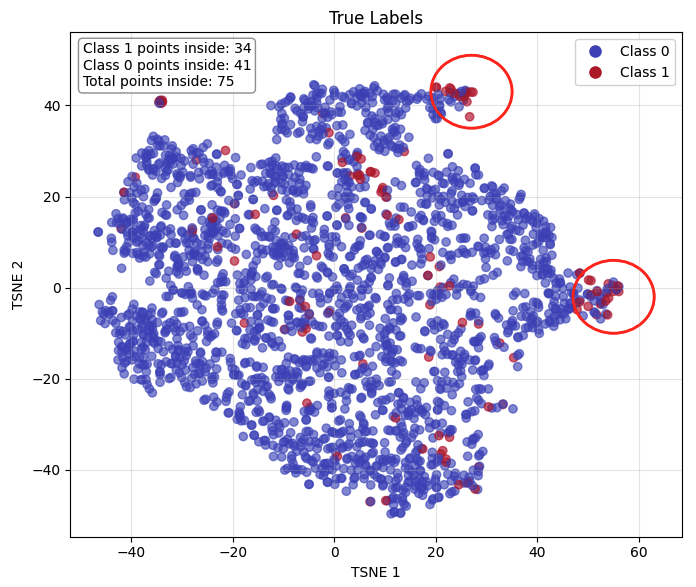}
  \caption{Embeddings of the transformer projected with t-SNE - Comparison between money laundering labels (class 0: Non-Fraudster and class 1: Fraudsters)}
  \label{fig:transformer_tsne_2}
\end{figure}
\begin{figure}[H]
  \centering
  \includegraphics[width=0.6\textwidth]{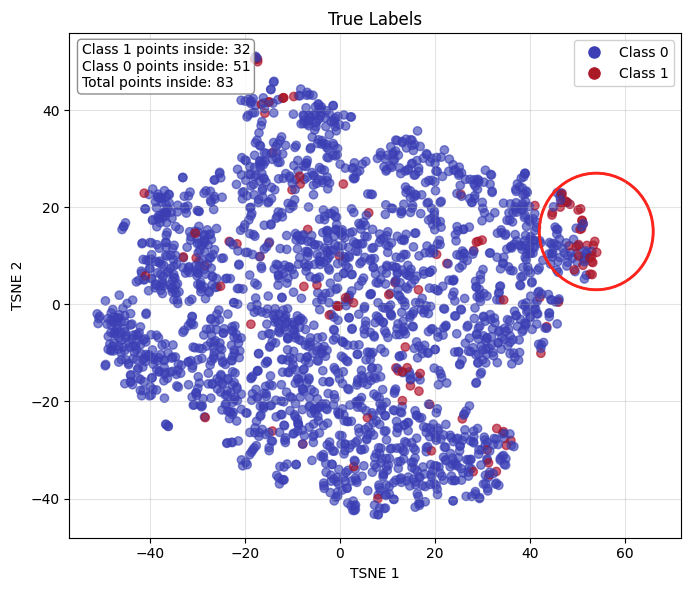}
  \caption{Embeddings of the transformer projected with t-SNE - Comparison between money laundering labels (class 0: Non-Fraudster and class 1: Fraudsters)}
  \label{fig:transformer_tsne_3}
\end{figure}
\begin{figure}[H]
  \centering
  \includegraphics[width=0.6\textwidth]{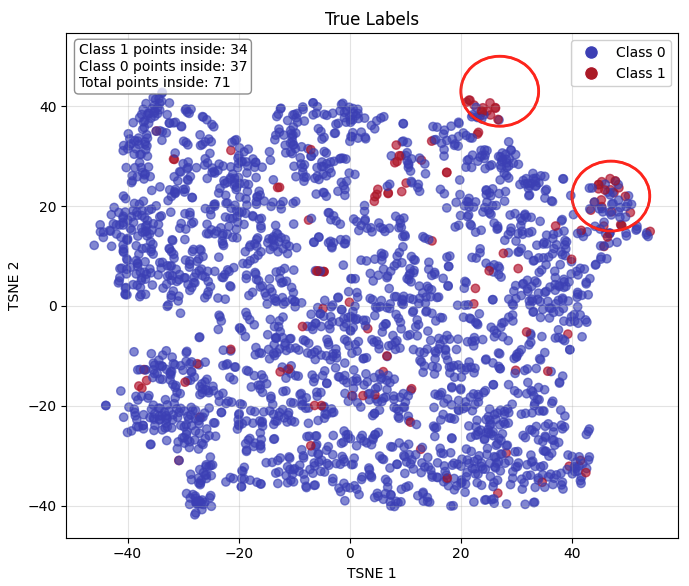}
  \caption{Embeddings of the transformer projected with t-SNE - Comparison between money laundering labels (class 0: Non-Fraudster and class 1: Fraudsters)}
  \label{fig:transformer_tsne_4}
\end{figure}

\subsubsection{LSTM Autoencoder}
\begin{figure}[H]
  \centering
  \includegraphics[width=0.6\textwidth]{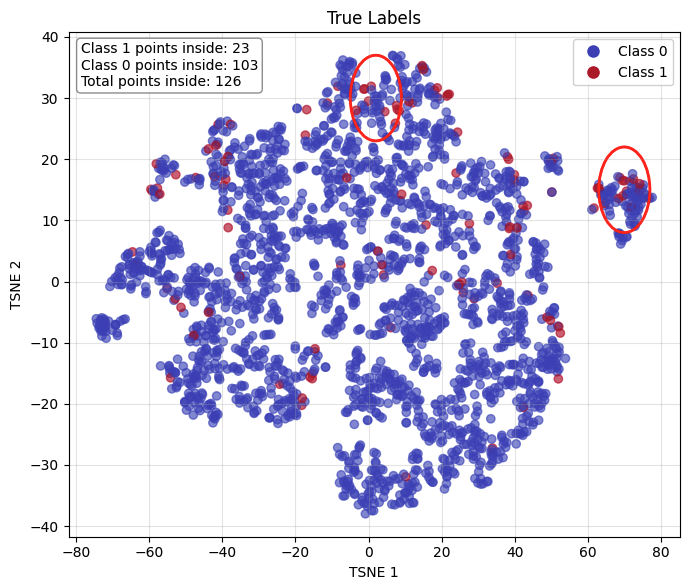}
  \caption{Embeddings of the LSTM Autoencoder projected with t-SNE - Comparison between money laundering labels (class 0: Non-Fraudster and class 1: Fraudsters)}
  \label{fig:lstm_tsne_2}
\end{figure}
\begin{figure}[H]
  \centering
  \includegraphics[width=0.6\textwidth]{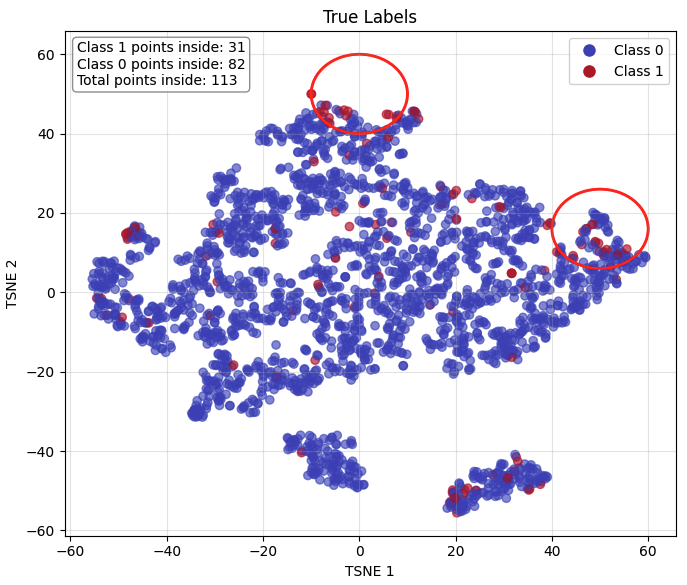}
  \caption{Embeddings of the LSTM Autoencoder projected with t-SNE - Comparison between money laundering labels (class 0: Non-Fraudster and class 1: Fraudsters)}
  \label{fig:lstm_tsne_3}
\end{figure}
\begin{figure}[H]
  \centering
  \includegraphics[width=0.6\textwidth]{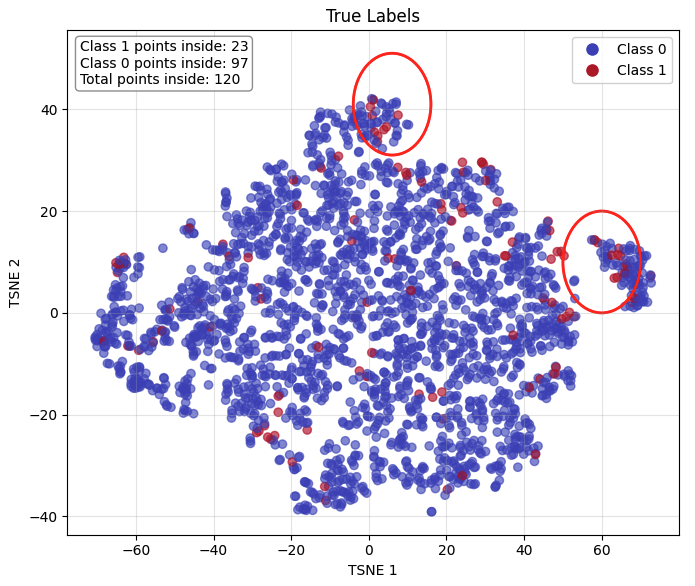}
  \caption{Embeddings of the LSTM Autoencoder projected with t-SNE - Comparison between money laundering labels (class 0: Non-Fraudster and class 1: Fraudsters)}
  \label{fig:lstm_tsne_4}
\end{figure}

\subsection{High thresholds - full tables}
\label{appendix.thresholds}
\begin{table}[H]
\begin{tabular}{lllll}
Model                                  & $FDR_h$ & Frauds detected & \% total frauds & F1-score \\
\hline
Logistic Regression - tabular data & 0.80 & 48 $\pm$ 0.0 & 0.41 & 0.27 \\
XGBoost - tabular data    & 0.80 & 43 $\pm$ 0.0 & 37.06 & 0.26 \\
Logistic Regression - LSTM & 0.80 & 49.7 $\pm$ 8.8 & 42.80 & 0.27 \\
\textbf{Logistic Regression - transformer} & \textbf{0.80} & \textbf{83.50 $\pm$ 2.4} & \textbf{72.0} & \textbf{0.31} \\
Transformer fine-tuning & 0.80 & 79.7 $\pm$ 3.4 & 68.70 & 0.30 \\
\hline
Logistic Regression - tabular data & 0.70 & 37.0 $\pm$ 0.0 & 31.90 & 0.31 \\
XGBoost - tabular data    & 0.70 & 22.0 $\pm$ 0.0 & 18.97 & 0.23 \\
Logistic Regression - LSTM & 0.70 & 37.7 $\pm$ 9.1 & 32.50 & 0.31 \\
\textbf{Logistic Regression - transformer} & \textbf{0.70} & \textbf{67.9 $\pm$ 5.5} & \textbf{58.53} & \textbf{0.39} \\
Transformer fine-tuning & 0.70 & 67.7 $\pm$ 4.9 & 58.36 & 0.39 \\
\hline
Logistic Regression - tabular data & 0.60 & 33.0 $\pm$ 0.0 & 28.45 & 0.33 \\
XGBoost - tabular data    & 0.60 & 10.0 $\pm$ 0.0 & 8.62 & 0.14 \\
Logistic Regression - LSTM & 0.60 & 25.3 $\pm$ 11.4 & 21.81 & 0.28 \\
Logistic Regression - transformer & 0.60 & 54.8 $\pm$ 3.6 & 47.24 & 0.43 \\
\textbf{Transformer fine-tuning} & \textbf{0.60} & \textbf{55.7 $\pm$ 3.4} & \textbf{48.02} & \textbf{0.43} \\
\hline
Logistic Regression - tabular data & 0.50 & 16.0 $\pm$ 0.0 & 13.79 & 0.22 \\
XGBoost - tabular data    & 0.50 & 1.0 $\pm$ 0.0 & 0.86 & 0.02 \\
Logistic Regression - LSTM & 0.50 & 16.4 $\pm$ 11.8 & 14.15 & 0.22 \\
Logistic Regression - transformer & 0.50 & 42.1 $\pm$ 5.4 & 36.29 & 0.42 \\
\textbf{Transformer fine-tuning} & \textbf{0.50} & \textbf{44.6 $\pm$ 6.2} & \textbf{38.45} & \textbf{0.43} \\
\hline
Logistic Regression - tabular data & 0.40 & 0.0 $\pm$ 0.0 & 0.0 & 0.0 \\
XGBoost - tabular data    & 0.40 & 1.0 $\pm$ 0.0 & 0.86 & 0.02 \\
Logistic Regression - LSTM & 0.40 & 8.5 $\pm$ 6.2 & 7.33 & 0.13 \\
Logistic Regression - transformer & 0.40 & 23.6 $\pm$ 12.8 & 20.34 & 0.30 \\
\textbf{Transformer fine-tuning} & \textbf{0.40} & \textbf{26.7 $\pm$ 14.5} & \textbf{23.02} & \textbf{0.33} \\
\hline
Logistic Regression - tabular data & 0.30 & 0.0 $\pm$ 0.0 & 0.0 & 0.0 \\
XGBoost - tabular data    & 0.30 &0.0 $\pm$ 0.0 & 0.0 & 0.0 \\
Logistic Regression - LSTM & 0.30 & 5.5 $\pm$ 4.0 & 4.74 & 0.089 \\
\textbf{Logistic Regression - transformer} & \textbf{0.30} & \textbf{12.3 $\pm$ 11.6} & \textbf{10.60} & \textbf{0.18} \\
Transformer fine-tuning & 0.30 & 10.2 $\pm$ 9.2 & 8.79 & 0.15 \\
\hline
Logistic Regression - tabular data & 0.20 & 0.0 $\pm$ 0.0 & 0.0 & 0.0 \\
XGBoost - tabular data    & 0.20 & 0.0 $\pm$ 0.0 & 0.0 & 0.0 \\
Logistic Regression - LSTM & 0.20 & 2.6 $\pm$ 1.9 & 2.41 & 0.04 \\
\textbf{Logistic Regression - transformer} & \textbf{0.20} & \textbf{4.0 $\pm$ 5.0} & \textbf{3.45} & \textbf{0.06} \\
Transformer fine-tuning & 0.20 & 0.8 $\pm$ 2.5 & 0.69 & 0.01 \\
\hline
Logistic Regression - tabular data & 0.10 & 0.0 $\pm$ 0.0 & 0.0 & 0.0 \\
XGBoost - tabular data    & 0.10 & 0.0 $\pm$ 0.0 & 0.0 & 0.0 \\
Logistic Regression - LSTM & 0.10 & 0.0 $\pm$ 0.0 & 0.0 & 0.0 \\
\textbf{Logistic Regression - transformer} & \textbf{0.10} & \textbf{0.9 $\pm$ 2.8} & \textbf{0.78} & \textbf{0.01} \\
Transformer fine-tuning & 0.10 & 0.0 $\pm$ 0.0 & 0.0 & 0.0 \\
\hline
\end{tabular}
\caption{Performance summary on the high threshold - Full table}
\label{tab:perf.summary.high.threshold.full}
\end{table}

\end{document}